\documentclass[1p]{elsarticle}

\usepackage{hyperref,tabularx,amsmath,amssymb,bm, mathtools}
\usepackage[list=true, font=normalsize, labelfont=sf,
labelformat=brace, position=top]{subcaption}
\usepackage{pgfplots}
\usepackage{tikz}
\usepackage[c1]{optidef}

\usepackage{xspace}

\newcommand{\optCog}{optCog\xspace}
\newcommand{\optTire}{optTire\xspace}
\newcommand{\Init}{Init\xspace}

\newcommand{\TireMod}{\ensuremath{\boldsymbol{T}}\xspace}
\newcommand{\PredMod}{\ensuremath{\boldsymbol{P}}\xspace}
\newcommand{\xoptCog}{\ensuremath{\dot{\boldsymbol{x}}^{\mathrm{oc}}}\xspace}
\newcommand{\xv}{\ensuremath{\dot{\boldsymbol{x}}}\xspace}
\newcommand{\xInit}{\ensuremath{\dot{\boldsymbol{x}}^{\mathrm{in}}}\xspace}
\newcommand{\yoptCog}{\ensuremath{\dot{\boldsymbol{y}}_{\mathrm{as}}^{\mathrm{oc}}}\xspace}
\newcommand{\yv}{\ensuremath{\dot{\boldsymbol{y}}}\xspace}
\newcommand{\yoptTire}{\ensuremath{\dot{\boldsymbol{y}}_{\mathrm{as}}^{\mathrm{ot}}}\xspace}

\newcommand{\SVeh}{\ensuremath{S_\mathrm{veh-traj}}\xspace}
\newcommand{\SHand}{\ensuremath{S_\mathrm{handling}}\xspace}

\newcommand{\STot}{\ensuremath{S_\mathrm{tot}}\xspace}

\newcommand{\nEngine}{\ensuremath{n_\mathrm{Engine}}\xspace}
\newcommand{\rThrottle}{\ensuremath{r_\mathrm{Throttle}}\xspace}
\newcommand{\pBrake}{\ensuremath{r_\mathrm{Brake}}\xspace}
\newcommand{\aRocker}{\ensuremath{r_\mathrm{Rocker,as}}\xspace}

\begin{document}


%
%
%
%



\journal{Vehicle System Dynamics}

\begin{frontmatter}
\title{Race Driver Evaluation at a Driving Simulator using a physical Model and a Machine Learning Approach\tnoteref{mytitlenote}}
\tnotetext[mytitlenote]{This work was generously supported by BMW AG.}

\author{Julian von Schleinitz\fnref{bmw} \corref{mycorrespondingauthor}}
\author{Thomas Schwarzhuber \fnref{bmw}}
\author{Lukas Wörle \fnref{bmw}}
\author{Michael Graf \fnref{ge}}
\author{Arno Eichberger \fnref{ug}}
\author{Wolfgang Trutschnig \fnref{us}}
\author{Andreas Schröder \fnref{us}}

\cortext[mycorrespondingauthor]{Corresponding author}
\fntext[bmw]{BMW AG}
\fntext[ge]{Graf Engineering}
\fntext[ug]{Technical University of Graz}
\fntext[us]{University of Salzburg}

\begin{abstract}
Professional race drivers are still superior  to automated systems at controlling a vehicle at its dynamic limit. Gaining insight into race drivers' vehicle handling process might lead to further development in the areas of automated driving systems. We present a method to study and evaluate race drivers on a driver-in-the-loop simulator by analysing tire grip potential exploitation. Given initial data from a simulator run, two optimiser based on physical models maximise the horizontal vehicle acceleration or the tire forces, respectively.
An overall performance score, a vehicle-trajectory score  and a handling score are introduced to  evaluate drivers.  Our method is thereby completely track independent and can be used from one single corner up to a large data set.   We apply the proposed method to a  motorsport data set containing over 1200 laps from seven professional race drivers and two amateur drivers whose lap times are $10-20\%$ slower. The difference  to the professional drivers comes mainly from their inferior handling skills and not their choice of driving line. 
A downside of the presented  method for certain applications is an extensive computation time. Therefore, we propose a Long-short-term memory (LSTM) neural network to estimate the driver evaluation scores. We show that the neural network is accurate and robust with a root-mean-square error between $2-5\%$ and can  replace the optimisation based method. The time for processing the data set considered in this work is reduced from $68$ hours to $12$ seconds, making the neural network suitable for real-time application.

\end{abstract}


\end{frontmatter}



\section{Introduction}
 With the advance of autonomous driving, self-driving cars will continue to encounter emergency situations \cite{Kegelman.2018}. In these situations the vehicle has to be controlled at its dynamic limit. A task at which most road car drivers are not trained at and lack the required experience. Expert race drivers in contrast exploit the vehicles capabilities to a much greater extent.
 Gaining insight in race drivers' vehicle handling process might therefore lead to further development  in the areas of automated driving, e.g. in case of critical scenarios at the vehicle's dynamic limit. Data from motorsport applications are ideally suited for that task since both engineers and race drivers spend a considerable amount of time on optimising the driver-vehicle system. However, most effort is put into developing the vehicle and not into analysing  drivers.\\
The aim of this work is to provide evaluation methods for drivers which allow general, track independent conclusions on the driving style and point out areas for performance improvement. The best conditions for analysing driving styles provides a driver-in-the-loop simulator since all parameters and circumstances are exactly known at every time step and the simulated environment guarantees reproducible conditions. We develop our method to evaluate race drivers based on the BMW Motorsport simulator which has been in use since 2017. Great effort has been put into matching the simulator with reality, which is the foundation of this work to analyse the performance of race drivers. 


\subsection{Background}
\label{chap_background}

\begin{figure}
\begin{center}
\includegraphics[width=0.7\textwidth]{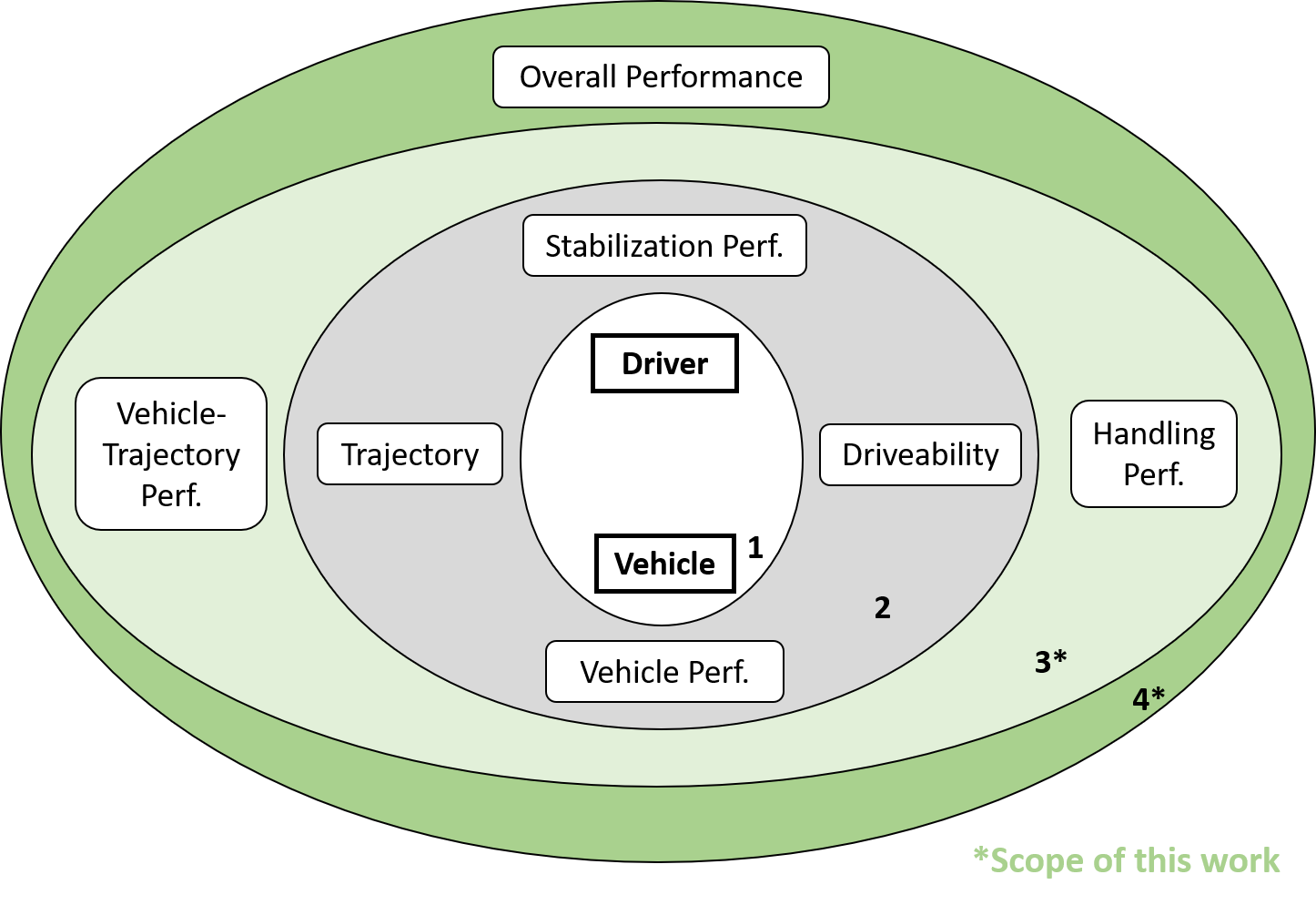}
\caption{Simplified model to describe how driver and vehicle interactions lead to the overall performance. The driver can choose a trajectory and has a stabilisation performance. The stabilisation performance with driveability lead to the total handling performance. In this work, we present a method to achieve the separation in handling and vehicle/trajectory performance in layer 3.}
\label{fig_perf_overview}
\end{center}
\end{figure}

The overall performance in motorsport is the result of driver-vehicle interaction. On a race track, overall performance is usually determined by the lap time. Given a desired trajectory around a race track, the lap time is minimised by maximising the average velocity. Two factors limit maximising the velocity. First, the vehicle can be restrained by the power of the engine to accelerate further. This is called 'power limited'. Second, there is  an acceleration limit for the vehicle due to tire-road friction. Track sections where maximising the velocity would violate this limit are referred to as 'grip limited'.
Since the power limitation cannot be overcome by the driver we focus on the grip limited track sections in this work. Along a defined trajectory, a vehicle can either accelerate or decelerate. Deceleration to a lower speed is necessary before grip limited track sections (usually corners) that otherwise would require  accelerations above the vehicle's capabilities to stay on the desired trajectory. In order to drive fast for as long as possible these deceleration phases need to be as short as possible and the amount of deceleration needs to be maximal. Thus, it is reasonable for the context of this work to define optimal performance as the maximum absolute horizontal acceleration achievable while retaining the spatial path of the vehicle's trajectory. In other words, for optimal performance the available grip should be completely exploited by the vehicle, it would then always be grip or power limited.  \\
The system driver-vehicle has to be described as a closed-loop system. In order to attempt a separation within this system we  divide it into four layers as presented in \autoref{fig_perf_overview}. Desirable would be a division into driver and vehicle related performance to determine specific areas of improvement. The \emph{overall performance}  in layer 4, as defined above, results from layer 3 which contains the combined \emph{vehicle-trajectory performance} as well as the \emph{handling performance} to keep the vehicle on this trajectory. \\
On layer 2 we have the \emph{stabilisation performance} to account for the driver's skills to stabilise the vehicle on a given trajectory and the \emph{vehicle performance} to capture the vehicle's capabilities. Influenced by both driver and vehicle are  \emph{trajectory} and \emph{driveability}. The trajectory results naturally from the driver's inputs and the vehicle's reaction. Depending on the vehicle setup different trajectories are more or less time-optimal. Driveability in the context of motorsport captures how feasible it is for a driver to extract the vehicle's potential.  
In this work we focus on layer 3 and 4, that is the overall, vehicle-trajectory and handling performance.

\subsection{Related work}

\citet{Donges.1982} developed a three level approach of the driving task from an engineering point of view. There are  the navigation level, guidance level and stabilisation level. The navigation level includes choosing the appropriate route from a given road infrastructure while also taking an approximate time needed for the route into account. For motorsport application, the course around a race track is predefined, therefore the navigation level is not relevant. The main driving process happens in the guidance and stabilisation level. The guidance level consists of deriving a target trajectory and target velocity, considering the planned driving route and the constantly changing conditions of the track environment. According to the derived targets, control actions are chosen in an anticipatory manner to yield the best possible initial condition with the least deviations from the targets. On the stabilisation level the driver has to manipulate the vehicle controls to keep the deviations to the target trajectory at a minimum level and interacts with the vehicle in a closed loop system. Thereby, the human driver's choice of actions does not solely depend on the current state, but also on prior knowledge and experience according to \citet{Macadam.2003}. 
In \cite{Worle.2019, Woerle.,Kegelman.2017b} it is discussed that race drivers have different driving styles while achieving similar lap times. 
\citet{Worle.2019} defined objective criteria for race drivers in professional motorsport, based on their driving control inputs and on their chosen driving trajectories. The algorithm automatically detects cornering manoeuvres and uses pattern recognition to classify different drivers. 
Similarly, \citet{Schleinitz.2019} found that top level race drivers could be distinguished with high accuracy by analysing only the brake and throttle pedal signals from data of a single corner. 
 \citet{Kegelman.2017b} presented a case study where two professional race drivers employed different driving styles to achieve similar lap times. Accordingly, this supports the idea that driving at the dynamic limit allows a family of solutions in terms of paths and speed. 
\citet{Segers.2014} evaluated driving behaviour of race car drivers, by introducing measures based on the performance, smoothness, response and consistency of the driver input signals. The measures were used to objectively detect differences between drivers.
\citet{Lockel.2020} proposed a probabilistic framework for imitating race drivers. They defined a set of metrics based on driver inputs, for example steering or braking aggressiveness to evaluate their framework. \\
Common motorsport metrics such as lap time, top speed or accelerations provide performance measures but do not give insights how they were achieved nor why certain traits are observed. Also properties which are linked to handling and therefore to driveability are difficult to extract from these metrics \cite{goy.16}.  \\
Autonomous race driving is an area where understanding of vehicle handling at the limit is crucial. 
\citet{Kegelman.2018} conducted an analysis of data from vintage race cars and quantified the dispersion of paths driven by drivers during races. An autonomous race car is used to research whether the variance observed in human driving is due to error or a consequence of purposefully  exploiting the vehicle limits. It showed that expert human drivers operated the vehicle around and also beyond the stability limit on purpose, whereas the autonomous vehicle always remained within the stable region. This way the expert drivers were significantly faster than the autonomous vehicle. 
\citet{Hermansdorfer.20.05.2020} compared an autonomous software stack to professional human race drivers using common motorsport analysis techniques and came to a similar conclusion. The autonomous software followed the trajectory smoothly but did not drive at the handling limit and was therefore inferior to human drivers. Both studies indicate that a very important talent of race drivers is constantly exploiting the available grip.

However, in reality this is hard to determine since the available grip changes constantly and with it the current dynamic limit of the vehicle. Here, driver-in-the-loop simulators have a big advantage for analysing drivers. 
\citet{S.deGroot.2011} studied the effect of training inexperienced race drivers with different levels of grip on a simulator. They measured classical motorsport performance factors such as lap time, full throttle percentage and did questionnaires to assess the drivers confidence. 
\citet{vanLeeuwen.2017} determined the differences between racing and non-racing drivers on a simulator using eye-tracking. They observed that race drivers showed a more variable gaze behaviour. \\
Studies about driver evaluation in terms of vehicle handling on a simulator are rare. A possible reason is that motorsport suited simulators are very expensive and  motorsport teams usually do not publish their findings. \citet{Schwarzhuber.2020}  introduced the TPER method (tire potential exploitation rating) whereby the difference between the vehicle's acceleration and an optimised acceleration based on a two-track vehicle and a Pacejka tire model is used as an evaluation criteria for different motion cueing algorithms. They investigated steady-state manoeuvres and concluded that TPER is a powerful tool for simulator fidelity investigations. \\
This work is based on the TPER method, but focuses on driver evaluation instead of simulator development. Therefore, not only steady-state manoeuvres, but complete laps are evaluated. In addition to optimising the vehicle's acceleration, we introduce four optimisers to determine the maximum force on each tire. Comparing the initial state to these optimised states  allows to calculate scores for overall, vehicle-trajectory and handling performance.

\subsection{Overview}
\label{chap_problem}
In chapter 2, the simulator and the vehicle model are detailed and the data set is presented. \\
Chapter 3 describes the methodology. 
We optimise the vehicle's acceleration in the center of gravity (Cog) for which we introduce the \optCog optimiser and compare it to the initial state. 
In order to evaluate the performance of the vehicle, a second optimiser which we named \optTire evaluates the maximum possible force of each tire independently. 
In the next step we define scores based on the outcome of the optimiser models that allow differentiation between overall, vehicle-trajectory and handling performance. 
We also propose a LSTM-based machine learning architecture to directly determine the scores form the initial data.\\ 
 Chapter 4 presents the results of this study. Having at hand the methods and the data set it is analysed how the drivers exploit the  grip by studying the slip ratios and slip angles at the tires and comparing them for the optimization states. 
We also compare the professional drivers to  amateur drivers. Lastly, the prediction performance of the machine learning model is shown.
Summarized, the main contributions of this work are:
\begin{itemize}
\item A method to evaluate race drivers on a simulator based on tire potential exploitation
\item Separation of the overall performance in handling and vehicle-trajectory performance
\item Application of the introduced method to a data set from professional race drivers on a top-level motorsport simulator
\item Comparison of professional drivers to two amateur drivers
\item A machine learning model architecture which can replace the optimization based method while being much faster to compute
\end{itemize}

\section{Apparatus}
The experiment to evaluate the methodology as an objective driver analysis tool is carried out at a four degrees of freedom (DOF) driver-in-the-loop simulator (DiLS). 

\subsection{Driver-in-the-loop simulator}
%
%

The DiLS used for the study is located in Munich and was designed by BMW Motorsport. Therefore, special focus has been put on its applicability for the motorsport environment. It has been operational since 2017. To this day, it is frequently used for vehicle development and driver training. 

The mechanical assembly consists of a static $210^{\mathrm{\circ}}$ curved screen. The screen surrounds a four DOF motion platform on top of which a mock-up of the corresponding race car's chassis is attached. Three of the motion platform's DOFs are rendering angular velocities: yaw rate $\dot{\psi}$, pitch rate $\dot{\theta}$, and roll rate $\dot{\phi}$. Translational motion is displayed in vertical direction (heave) only by means of the heave acceleration $a_{\mathrm{z}}$. The range and dynamic capabilities of each DOF were identified using classical system identification techniques. 
The motion platform is controlled by a custom motion cueing algorithm (MCA) \citep{Schwarzhuber.2020}. In addition to the motion and visualization system, stimuli are provided by means of a static sound system and the steering force feedback. Both provide relevant information about vehicle states, making them vital parts of the DiLS \cite{Liu.1995, Toffin.2003}. The steering wheel is directly driven by an electric motor which allows to display up to $24\,\mathrm{Nm}$ of steering torque. The torque demand is derived from the tire forces at the front axle and the vehicle's steering geometry.

All the above mentioned systems and their respective characteristics result in a certain simulator fidelity, which was proven to have an impact on driver-vehicle interaction \cite{Markkula.2019}. For the proposed methodology, the race drivers' behaviour is a fundamental part. In the context of driving simulation the term of  validity refers to similar driving behaviour between simulator and in the real world driving tasks. An earlier study conducted at the BMW Motorsport DiLS quantitatively identified its validity \cite{Schwarzhuber.2020b}. Detailed results are not relevant for the present work but should be considered if an extension to real driving tasks is sought.

\subsection{Vehicle Model}
The vehicle and tire models are described separately as these are fundamental parts not only of the driving simulator but also of the methodology itself. For vehicle dynamics modelling, a custom two track model is implemented similar as described in \cite{Heiing.2013}. The proposed methodology relies on detailed knowledge of the models. It is a mandatory requirement to have all vehicle and tire model parameters available in order to make the methodology feasible. The models' inputs are twofold. Firstly, the driver controls the vehicle via steering wheel including gear-shifter functionality, the accelerator pedal, brake pedal and clutch pedal. Secondly, a terrain server provides road inputs by means of tires' contact patch coordinates. In order to further process contact patch and tire load information to tire forces, the simulation model is equipped with a Pacejka Magic Formula tire model \cite{Pacejka.1992}. The parameters of vehicle and tire models are identified individually by means of test bench data. Subsequently, the models are validated using mainly closed loop manoeuvres. Validity in this context is only defined for circuit racing close to the vehicle's maximum capabilities. The model in use was confirmed to provide relative validity, meaning that vehicle setup variations show the same sensitivities in the virtual and the real environment. 

For the purpose of driving simulation, all models are required to be realtime executable. Deploying a compiled version of the models to a \emph{Speedgoat} real-time target machine, allows an execution frequency of $2$ kHz.

\subsection{Data}
\begin{figure}
\begin{center}
\includegraphics[width=0.7\textwidth]{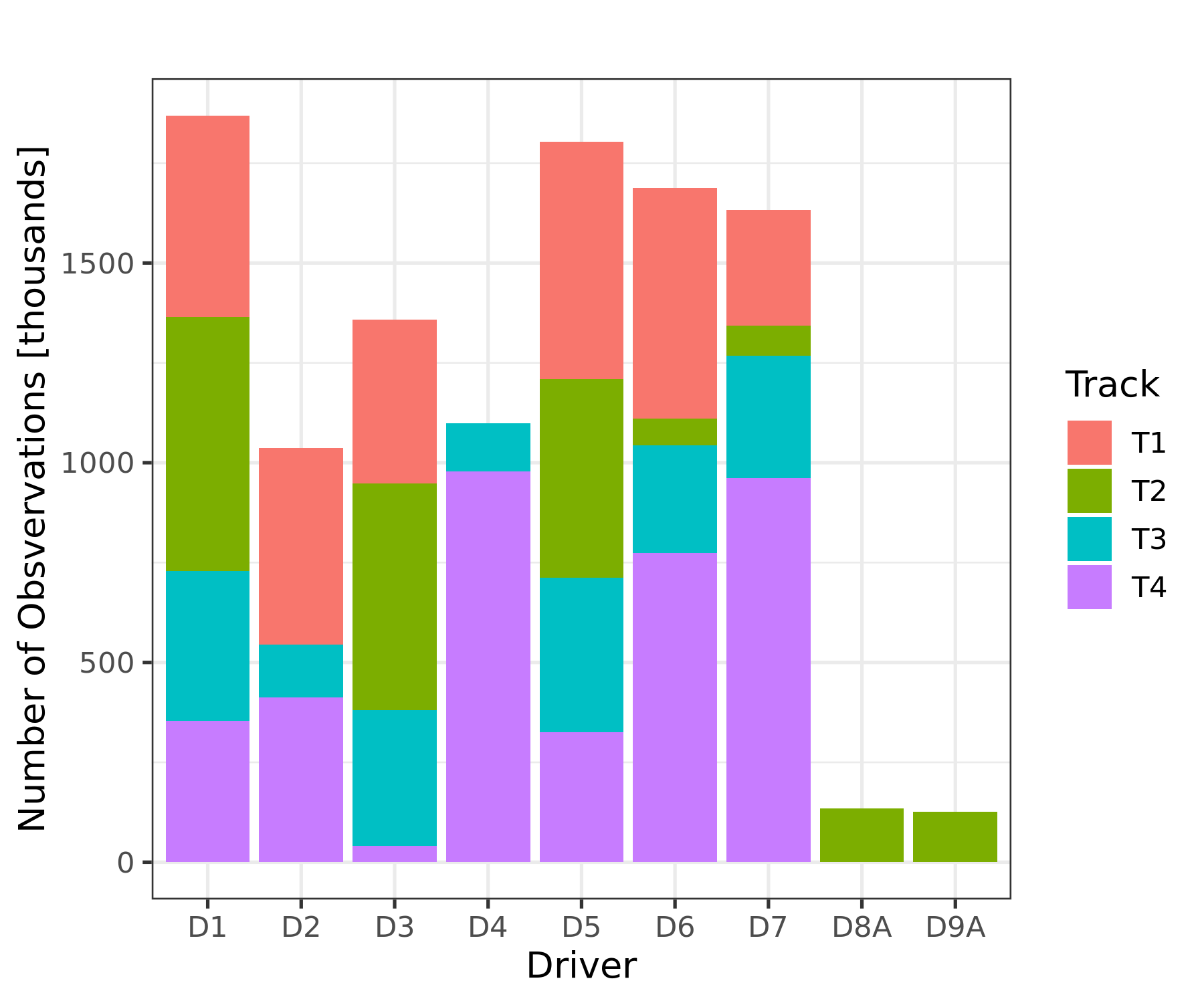}
\caption{Overview of the data set used in this work. The two amateur drivers D8A and D9A only drove on one race track. In total there are over $10$ million observations with a sample rate of $100$ Hz.}
\label{fig_data}
\end{center}
\end{figure}

The basis of this work is a data set from BMW motorsport. It consists of time series signals from a professional
motorsport racing series which are obtained from the aforementioned simulator. 
\autoref{fig_data} provides an overview of the data set. It was logged with a sample rate of $100$ Hz and there are over $10$ million instances. The data were collected in 2020 as a part of race preparation and consist of seven drivers (D1-D7) on four tracks (T1-T4). Additionally, data from two amateur drivers (D8A and D9A) were collected on track T2. 

\section{Methods}

\begin{figure}
\begin{center}
\includegraphics[width=0.7\textwidth]{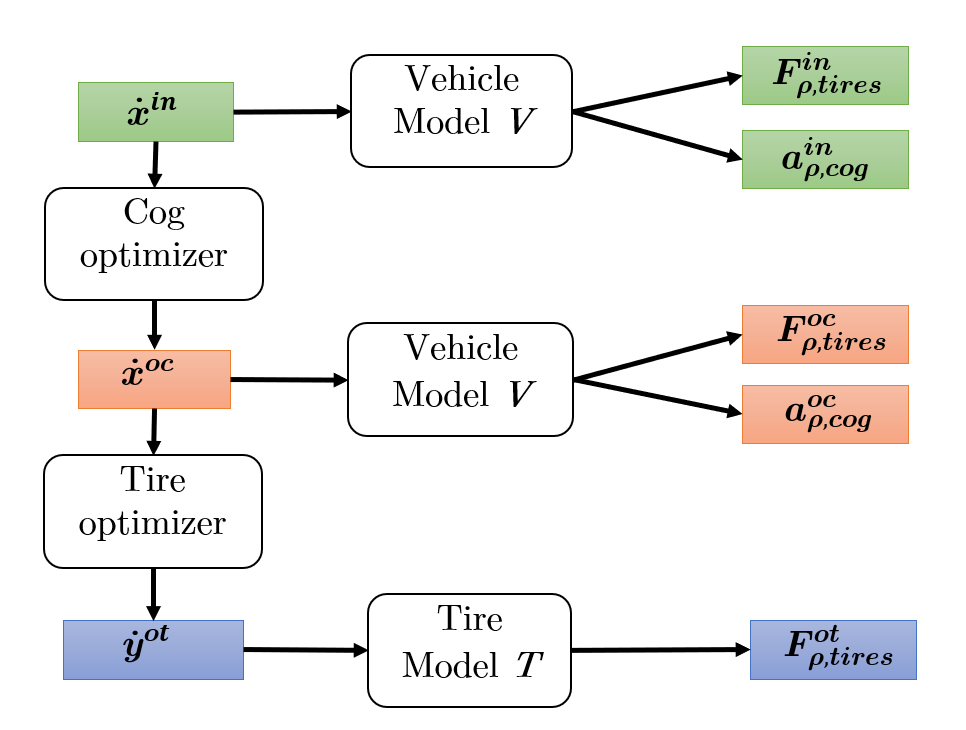}
\caption{Path from the \Init state \xInit over the \optCog state \xoptCog to the \optTire state \yoptTire. Using a vehicle or tire model forces $F$ and accelerations $a$ can be calculated from the state vectors $\dot{x}$ and $\dot{y}$. }
\label{fig_optimization_states}
\end{center}
\end{figure}

\subsection{Dynamic limit optimisation}
The purpose of the method described in this section is to evaluate how far the driver or the vehicle is  from the optimum, which is given by the maximum horizontal acceleration while retaining the trajectory or having the highest possible tire forces, respectively. We split the driving task into two parts. According to the three-level model of driving tasks choosing a trajectory corresponds to the guidance level and stabilising the vehicle along this trajectory to the stabilisation level \cite{Worle.20,Donges.1982}.

\subsubsection{optCog optimisation}
\begin{figure}
\begin{center}
\begin{subfigure}[c]{0.7\textwidth}
\begin{tikzpicture}[scale=0.85]

\draw [line width = 1mm, color={rgb:red,1.3;green,1.8;blue,2.1}] plot[smooth, tension=.7] coordinates {(-5.7,-2.55) (-3.5,-0.5) (-0.5,0.5) (4.65,0.5)};
\draw  [line width = 1mm, color={rgb:red,1.3;green,1.8;blue,2.1}] plot[smooth, tension=.7] coordinates {(-4,-4.6) (-2.1,-2.7) (0,-2) (4.6,-2)};
\draw  [dashed] plot[smooth, tension=.7] coordinates {(-5.05,-3.45) (-4.3,-2.85) (-3.05,-2.05) (-0.8,-1.05) (1.95,-0.45) (4.55,-0.1)};

\draw  (-2.75,-1.9) ellipse (0.25 and 0.25);
\draw  (1.55,-0.5) ellipse (0.25 and 0.25);

\draw [line width = 0.3mm, color=gray] (0.7,-0.5) -- (2.7,-0.5);
\draw [line width = 0.3mm, color=gray] (0.7,0) -- (0.7,-1);
\draw [line width = 0.3mm, color=gray] (2.7,0) -- (2.7,-1);
\draw  [fill, color=gray] (0.3,0.25) rectangle (1.1,0);
\draw  [fill, color=gray] (0.3,-1) rectangle (1.1,-1.25);
\draw  [fill, color=gray] (2.3,0.25) rectangle (3.1,0);
\draw  [fill, color=gray] (2.3,-1) rectangle (3.1,-1.25);

\draw [line width = 0.3mm, color=gray] (-1.7,-1.3) -- (-3.55,-2.35);
\draw [line width = 0.3mm, color=gray] (-2,-0.85) -- (-1.4,-1.75);
\draw [line width = 0.3mm, color=gray] (-3.85,-1.9) -- (-3.25,-2.8);

\draw [fill, color=gray] (-2.47,-0.67) -- (-1.7,-0.53) -- (-1.65,-0.77) -- (-2.4,-0.92) -- (-2.47,-0.67);
\draw [fill, color=gray] (-1.82,-1.82) -- (-1.05,-1.66) -- (-1,-1.9) -- (-1.75,-2.05) -- (-1.82,-1.82);
\draw [fill, color=gray] (-3.65,-3) -- (-2.9,-2.55) -- (-2.75,-2.75) -- (-3.5,-3.2) -- (-3.65,-3);
\draw [fill, color=gray] (-4.35,-1.9) -- (-3.6,-1.45) -- (-3.45,-1.65) -- (-4.2,-2.1) -- (-4.35,-1.9);

\filldraw[fill=black] (-2.75,-1.9) -- (-2.5,-1.9) arc (0:90:0.25cm) -- cycle;
\filldraw[fill=black] (-2.75,-1.9) node (v1) {} -- (-3,-1.9) arc (0:90:-0.25cm) -- cycle;

\filldraw[fill=black] (1.55,-0.5) -- (1.3,-0.5) arc (0:90:-0.25cm) -- cycle;
\filldraw[fill=black] (1.55,-0.5) node (v2) {} -- (1.8,-0.5) arc (0:90:0.25cm) -- cycle;

\draw [line width = 0.4mm] (-2.75,-1.9) node (v3) {} -- (-1.5,-3) [-latex] node [below=1mm, left] {$a^{\mathrm{in}}_{\mathrm{cog}}$};
\draw [line width = 0.4mm] (1.55,-0.5) node (v4) {} -- (4,-1) [-latex] node [below = 2mm, left = -0.1mm] {$a^{\mathrm{in}}_{\mathrm{cog}}$};
\draw [line width = 0.2mm, color=red] (-2.75,-1.9) -- (-1.15,-3.3) [-latex] node [below = 3mm, left = -3mm] {$a^{\mathrm{oc}}_{\mathrm{cog}}$};
\draw [line width = 0.2mm, color=red] (1.55,-0.5) -- (4.6,-1.12) [-latex] node [below = 3mm, left = -2mm] {$a^{\mathrm{oc}}_{\mathrm{cog}}$};

\node at (-2.7,-4.5) {$\dot{\bold{x}}(k)$};
\draw (-2.15,-4.5) -- (-0.9,-4.5) node (v5) {};
\draw [dotted, line width = 0.3mm] (-0.9,-4.5) -- (-0.1,-4.5) node (v6) {};
\draw (-0.1,-4.5) -- (0.75,-4.5) [-latex];
\node at (1.5,-4.5) {$\dot{\bold{x}}(k+n)$};

\draw [dotted, line width = 0.3mm] (-0.5,0.7) -- (-0.5,-3.5);

\draw (-4.78,-3.68) node [right = 7mm, above = 1mm] {s} -- (-4.2,-3.2) [-latex];

\draw (-4.85,-3.6) -- (-4.7,-3.75);
\end{tikzpicture}

\subcaption{Visualisation of the optCog optimisation results for two samples. The amount of acceleration in the center of gravity is maximised. The optimisation keeps the initial vehicle trajectory.  }
\label{fig:TPERvehicle}
\end{subfigure}

\begin{subfigure}[c]{0.5\textwidth}
\begin{tikzpicture}[scale=1.7]

\draw [line width = 1mm, color={rgb:red,1.3;green,1.8;blue,2.1}] plot[smooth, tension=.7] coordinates {(-5.5668,-2.2503) (-3.5,-0.5) (-1.0994,0.3002)};
\draw  [line width = 1mm, color={rgb:red,1.3;green,1.8;blue,2.1}] plot[smooth, tension=.7] coordinates {(-3.22,-4.5061) (-1.9824,-3.435) (-0.5406,-2.9348)};

\draw  (-2.75,-1.9) ellipse (0.25 and 0.25);

\draw [line width = 0.3mm, color=gray] (-1.7,-1.3) -- (-3.55,-2.35);
\draw [line width = 0.3mm, color=gray] (-2,-0.85) -- (-1.4,-1.75);
\draw [line width = 0.3mm, color=gray] (-3.85,-1.9) -- (-3.25,-2.8);

\draw [fill, color=gray] (-2.47,-0.67) -- (-1.7,-0.53) -- (-1.65,-0.77) -- (-2.4,-0.92) -- (-2.47,-0.67);
\draw [fill, color=gray] (-1.82,-1.82) -- (-1.05,-1.66) -- (-1,-1.9) -- (-1.75,-2.05) -- (-1.82,-1.82);
\draw [fill, color=gray] (-3.65,-3) -- (-2.9,-2.55) -- (-2.75,-2.75) -- (-3.5,-3.2) -- (-3.65,-3);
\draw [fill, color=gray] (-4.35,-1.9) -- (-3.6,-1.45) -- (-3.45,-1.65) -- (-4.2,-2.1) -- (-4.35,-1.9);

\filldraw[fill=black] (-2.75,-1.9) -- (-2.5,-1.9) arc (0:90:0.25cm) -- cycle;
\filldraw[fill=black] (-2.75,-1.9) node (v1) {} -- (-3,-1.9) arc (0:90:-0.25cm) -- cycle;

\draw [line width = 0.4mm, color=red] (-2.75,-1.9) node (v3) {} -- (-1.5,-3) [-latex] node [below=1mm, left] {$a^{\mathrm{oc}}_{\mathrm{cog}}$};


\node at (-2.406,-4.2354) {$\dot{\bold{x}}(k)$};

\draw (-4.6468,-3.3803) node [right = 7mm, above = 1mm] {s} -- (-4.0668,-2.9003) [-latex];

\draw (-4.7168,-3.3003) -- (-4.5668,-3.4503);

\draw [line width = 0.3mm, color=red](-2.0804,-0.7418) -- (-2.0205,-1.0759)[-latex] node [left] {$F^{\mathrm{oc}}_{\mathrm{FL}}$};
\draw [line width = 0.2mm, color=green](-2.0804,-0.7418) -- (-1.9557,-1.3189)[-latex] node [below=1mm, left] {$F^{\mathrm{ot}}_{\mathrm{FL}}$};

\draw [line width = 0.3mm, color=red](-1.3748,-1.8888) -- (-1.3149,-2.2229)[-latex] node [left] {$F^{\mathrm{oc}}_{\mathrm{FR}}$};
\draw [line width = 0.2mm, color=green](-1.3748,-1.8888) -- (-1.2501,-2.4659)[-latex] node [below=2mm,left=-1mm] {$F^{\mathrm{ot}}_{\mathrm{FR}}$};

\draw [line width = 0.3mm, color=red](-3.9334,-1.8002) -- (-3.4329,-1.8991)[-latex] node [below=-0.3mm] {$F^{\mathrm{oc}}_{\mathrm{RL}}$};
\draw [line width = 0.2mm, color=green](-3.9334,-1.8002) -- (-3.1329,-1.9657)[-latex] node [above=4mm, left=-3mm] {$F^{\mathrm{ot}}_{\mathrm{RL}}$};

\draw [line width = 0.3mm, color=red](-3.1984,-2.8884) -- (-2.6979,-2.9873)[-latex] node [left=5mm, below=-1mm] {$F^{\mathrm{oc}}_{\mathrm{RR}}$};
\draw [line width = 0.2mm, color=green](-3.1984,-2.8884) -- (-2.3979,-3.0539)[-latex] node [below] {$F^{\mathrm{ot}}_{\mathrm{RR}}$};

\draw  [fill=black] (-3.9412,-1.7942) ellipse (0.04 and 0.04);
\draw  [fill=black] (-3.206,-2.8826) ellipse (0.04 and 0.04);
\draw  [fill=black] (-2.0882,-0.7352) ellipse (0.04 and 0.04);
\draw  [fill=black] (-1.3822,-1.8824) ellipse (0.04 and 0.04);
\end{tikzpicture}

\subcaption{Qualitative visualisation of the optTire result. The amount of force in each tire is optimised individually. In general, the optimisation does not result in a valid vehicle state nor does it retain the initial trajectory.}

\end{subfigure}
\end{center}
\caption{Overview of both optimisation methods. They aim at providing a measure of how close the vehicle is to the optimum given the defined constraints. }
\label{fig:TPERcombined}
\end{figure}
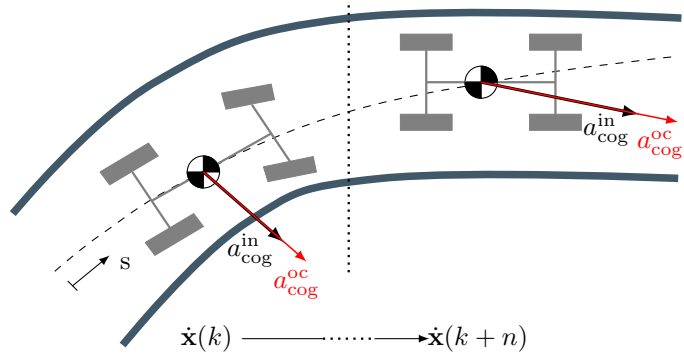
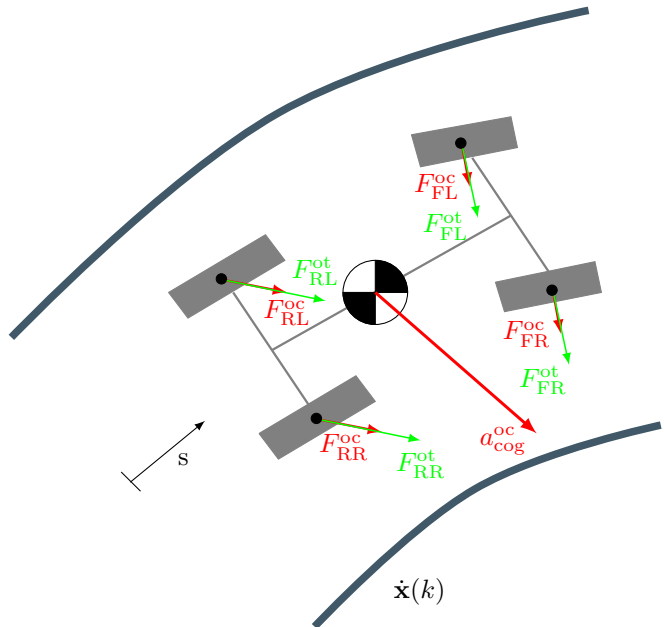

The \optCog optimisation is equivalent to the \textit{TPER} method by \citet{Schwarzhuber.2020}, we provide a brief summary of the method in this section. \autoref{fig:TPERvehicle} shows a graphical overview.
\textit{TPER}, which stands for Tire Potential Exploitation Rating, is based on a two track vehicle model as described in \cite{Jazar.2017} combined with a Pacejka Magic Formula tire model \cite{Pacejka.1992}.  The vehicle state vector \xInit is composed according to \eqref{eq:xIn}, where the index as with 'a'   $\in \{ \mathrm{f,r} \}$ and 's' $ \in \{ \mathrm{l,r}\}$ associates the corresponding wheel from front-left to rear-right on the vehicle. For simplicity the vehicle state vector is subdivided into a constant and a variable part, which changes during  optimisation
\begin{equation}
\begin{aligned}
 \xInit =[ &\xInit_{const} ~ \xInit_{var}]^{\textrm{T}}, \text{ with}\\
    \xInit_{const} =[&\dot{\psi}~v~Fz_\mathrm{as}~\alpha_\mathrm{as,{n-1}}~\mu_\mathrm{as}~b_\mathrm{as}~D_{\mathrm{x}}~D_{\mathrm{y}}~\\
    &\gamma_\mathrm{as}~r_\mathrm{as}~n_\mathrm{Engine}~i_\mathrm{Tot}~],\\
    \xInit_{var} =[ &\delta^\mathrm{in}~\beta^\mathrm{in}~\kappa_\mathrm{as}^\mathrm{in}].
\end{aligned}
	\label{eq:xIn}
\end{equation}
\autoref{tab:symbols} provides an overview of the parameters. The aim of the \optCog optimisation is to determine an optimised vehicle state vector from \xInit
\begin{equation}
\begin{aligned}
	\xoptCog =[ &\xInit_\mathrm{const} ~ \xoptCog_\mathrm{var}]^{\textrm{T}},\\
	\text{with } \xoptCog_\mathrm{var} =[ &\delta^\mathrm{oc}~\beta^\mathrm{oc}~\kappa^\mathrm{oc}_\mathrm{as}], \\
	\text{such that } L^{\mathrm{oc}}(\xoptCog )= & \min L^{\mathrm{oc}}(\dot{\bm{x}} ),
	\end{aligned}
	\label{eq:xOpt}
\end{equation}
whereby $L^{\mathrm{oc}}$ is the loss function and  $\xoptCog_\mathrm{var}$ are the optimisation variables. In consideration of the external forces $F_\mathrm{x,ext}$ and $F_\mathrm{y,ext}$, the longitudinal, lateral and angular momentum equalities of the vehicle model are defined. 
\begin{equation}
	\begin{aligned}
m \cdot a_\mathrm{x,cog}~=~&F_\mathrm{x,rl} \cdot \cos\delta_\mathrm{rl}+F_\mathrm{x,rr} \cdot \cos\delta_\mathrm{rr}-F_\mathrm{y,rl} \cdot \sin\delta_\mathrm{rl}- 
    F_\mathrm{y,rr} \cdot \sin\delta_\mathrm{rr}+ \\
    &F_\mathrm{x,fl} \cdot \cos\delta_\mathrm{fl}+F_\mathrm{x,fr} \cdot \cos\delta_\mathrm{fr}-
    F_\mathrm{y,fl} \cdot \sin\delta_\mathrm{fl}-F_\mathrm{y,fr} \cdot \sin\delta_\mathrm{fr}+ \\ &F_\mathrm{x,Ext}
    \end{aligned}
    \label{eq:ImpulseLong}
\end{equation}
\begin{equation}
\begin{aligned}
m \cdot a_\mathrm{y,cog}~=~&F_\mathrm{x,rl} \cdot \sin\delta_\mathrm{rl}+F_\mathrm{x,rr} \cdot \sin\delta_\mathrm{rr}+F_\mathrm{y,rl} \cdot \cos\delta_\mathrm{rl}+F_\mathrm{y,rr} \cdot \cos\delta_\mathrm{rr}+ \\ &F_\mathrm{x,fl} \cdot \sin\delta_\mathrm{fl} + F_\mathrm{x,fr} \cdot \sin\delta_\mathrm{fr}+F_\mathrm{y,fl} \cdot \cos\delta_\mathrm{fl}+F_\mathrm{y,fr} \cdot \cos\delta_\mathrm{fr}+ \\ &F_\mathrm{y,Ext}
\end{aligned}
\label{eq:ImpulseLat}
\end{equation}
\begin{equation}
\begin{aligned}
J_\mathrm{zz} \cdot \ddot{\psi}~=~&l_\mathrm{F} \cdot (F_\mathrm{x,fl} \cdot \sin\delta_\mathrm{fl}+F_\mathrm{x,fr} \cdot \sin\delta_\mathrm{fr}+F_\mathrm{y,fl} \cdot \cos\delta_\mathrm{fl}+F_\mathrm{y,fr} \cdot \cos\delta_\mathrm{fr})- \\ &l_\mathrm{R} \cdot (F_\mathrm{x,rl} \cdot \sin\delta_\mathrm{rl}+F_\mathrm{x,rr} \cdot \sin\delta_\mathrm{rr}+F_\mathrm{y,rl} \cdot \cos\delta_\mathrm{rl}+F_\mathrm{y,rr} \cdot \cos\delta_\mathrm{rr})+ \\ &b_\mathrm{fr} \cdot (F_\mathrm{x,fr} \cdot \cos\delta_\mathrm{fr}-F_\mathrm{y,fr} \cdot \sin\delta_\mathrm{fr})+ \\ &b_\mathrm{fl} \cdot (-F_\mathrm{x,fl} \cdot \cos\delta_\mathrm{fl}+F_\mathrm{y,fl} \cdot \sin\delta_\mathrm{fl})+ \\ &b_\mathrm{rr} \cdot (F_\mathrm{x,rr} \cdot \cos\delta_\mathrm{rr}-F_\mathrm{y,rr} \cdot \sin\delta_\mathrm{rr})+ \\ &b_\mathrm{rl} \cdot (-F_\mathrm{x,rl} \cdot \cos\delta_\mathrm{rl}+F_\mathrm{y,rl} \cdot \sin\delta_\mathrm{rl})+ \\ &M_\mathrm{z,fl}+M_\mathrm{z,fr}+M_\mathrm{z,rl}+M_\mathrm{z,rr}
\end{aligned}
\label{eq:ImpulseYaw}
\end{equation}
Equations \ref{eq:ImpulseLong} to \ref{eq:ImpulseYaw} enable to search for an optimal vehicle state vector \xoptCog in the sense of 
\begin{equation}
	L^{\mathrm{oc}}\left(\xv\right)~=~-\sqrt{(a_\mathrm{x,cog})^{2}+(a_\mathrm{y,cog})^{2}}.
	\label{eq:costfun}
\end{equation}
The objective of the optimisation is to maximise the norm of combined longitudinal ($a_{\mathrm{x,cog}}$) and lateral acceleration ($a_{\mathrm{y,cog}}$). The loss function $L^{\mathrm{oc}}\left(\xv\right)$ is formulated as a minimisation problem. 


\begin{table}[t]
	\centering 
	\caption{Description of the symbols used in this paper, whereby the index as with 'a' $\in \{ \mathrm{f,r} \}$ and 's' $ \in \{ \mathrm{l,r}\}$ associates the corresponding wheel from front-left to rear-right on the vehicle.}
	\small
	\begin{tabular}{l l}
		\hline	
		 \multicolumn{1}{c}{\bfseries Symbol} & \multicolumn{1}{c}{\bfseries Description} \\ \hline
		
		$\dot{\psi}$ & angular velocity around the vertical axis \\ \hline
		$v$ & resultant velocity of the vehicle \\ \hline
		$Fz_\mathrm{as}$ & tire loads \\ \hline
		$\alpha_\mathrm{as,{n-1}}$ & tire slip angles of the previous sample \\ \hline
		$\mu_\mathrm{as}$ & dynamic toe angles \\ \hline
		 $b_\mathrm{as}$ & half track widths \\ \hline
		  $D_{\mathrm{x}}$, $D_{\mathrm{y}}$ & longitudinal, lateral aerodynamic drag force \\ \hline
		    $\gamma_\mathrm{as}$ & camber angles \\ \hline
		     $r_\mathrm{as}$ & dynamic tire radii \\ \hline
		      $\nEngine$ & rotational speed of the engine \\ \hline
		       $i_\mathrm{Tot}$ & overall gearing from the engine to the tires\\ \hline
		           $\delta$ & steering angle \\ \hline
		            $\beta$ & body sideslip angle \\ \hline
		             $\kappa_\mathrm{as}$ & tires' longitudinal slip ratios \\ \hline
		             \xv & vehicle state vector \\ \hline
		             \yv & tire state vector \\ \hline
		             $a_\mathrm{x,cog},a_\mathrm{y,cog}$ & vehilce acceleration the center of gravity \\ \hline
		             $br_{\mathrm{dist}}$ & brake balance \\ \hline
		              $F_\mathrm{x,as},F_\mathrm{y,as}$ & forces at the tire contact patch \\ \hline
		              $\rThrottle$ & throttle pedal actuation \\ \hline
		              $\pBrake$ & brake pedal actuation \\ \hline
		              $\aRocker$ & rocker angle (suspension movement) \\ \hline
	\end{tabular}
	\label{tab:symbols}
\end{table}

For simplicity, the slip ratios are optimised directly, instead of modelling the entire powertrain and braking system. Though, powertrain limitations are formulated as constraints.  The first constraint  ensures driving trajectory compliance by keeping the approximated resultant angular acceleration $\ddot{\psi}$ constant
\begin{equation} 
	\ddot{\psi}^\mathrm{oc}-\ddot{\psi}^\mathrm{in}~=~0.
\label{eq:constraintsYawAcceleration}
\end{equation}

Furthermore, the direction of the resultant horizontal acceleration has to remain constant   
\begin{equation} 
	\dfrac{a_\mathrm{x,cog}^\mathrm{oc}}{a_\mathrm{y,cog}^\mathrm{oc}}-\dfrac{a_\mathrm{x,cog}^\mathrm{in}}{a_\mathrm{y,cog}^\mathrm{in}}~=~0.
\label{eq:constraintsARatio}
\end{equation}

Next, constraints for the slip ratios $\kappa_{\mathrm{as}}$ are introduced to consider the dependencies in the  vehicle model. The difference in driving and braking torque $T_{\mathrm{as}}$ between left and right tire has to remain constant on each axle  
\begin{equation} 
T^{\mathrm{oc}}_\mathrm{al}-T^{\mathrm{oc}}_\mathrm{ar}-(T^{\mathrm{in}}_\mathrm{al}-T^{\mathrm{in}}_\mathrm{ar})~=~0.
\label{eq:constraintsTorqueBraking}
\end{equation}

 Since only rear wheel driven cars are considered, the front axle constraint refers to braking torque only. The initial distribution of braking torques between front and rear axle has to remain unchanged. Therefore, a continuously differentiable braking distribution ($br_\mathrm{dist}$) constraint is formulated 
 \begin{equation} 
\begin{split}
&T_\mathrm{f}^{\mathrm{oc}}-(T_\mathrm{f}^{\mathrm{oc}}+T_\mathrm{r}^{\mathrm{oc}}) \cdot br_\mathrm{dist}~=~0,~\mathrm{with} \\
br_\mathrm{dist} ~\approx~ &br_\mathrm{drv} \cdot \left(\dfrac{\arctan(-\epsilon_\mathrm{1} \cdot (T_\mathrm{f}^{\mathrm{oc}}-\epsilon_\mathrm{2}))}{\pi}+\dfrac{1}{2}\right).
\end{split}
\label{eq:constraintsBrakingBias}
\end{equation}

The variable $br_\mathrm{drv}$ is the constant braking distribution selected by the driver and $\epsilon_1$ and $ \epsilon_2$ are small numerical parameters.  The range of possible side slip angles and  slip ratios is bound to
\begin{equation} 
\left(\dfrac{\alpha^{\mathrm{oc}}_\mathrm{as}}{\alpha_\mathrm{max}}\right)^{2}-1~\leq~0,
\label{eq:ineqConstraintsSideSlipAngle}
\end{equation}
\begin{equation} 
\left(\dfrac{\kappa^{\mathrm{oc}}_\mathrm{as}}{\kappa_\mathrm{max}}\right)^{2}-1~\leq~0.
\label{eq:ineqConstraintsSlipRatio}
\end{equation}
  
 Lastly, the maximum driving torque $T_\mathrm{r}$ is limited to the engine capabilities $T_{\mathrm{Engine}_{\mathrm{max}}}$ by 

\begin{equation} 
T_\mathrm{r}^{\mathrm{oc}}-T_\mathrm{Engine_{max}} \cdot i_\mathrm{Tot}~\leq~0.
\label{eq:ineqConstraintsTorqueMax}
\end{equation}

The optimization is carried out with the \textit{fmincon} solver of the \textit{MathWorks$^{\textrm{\textregistered}}$ MATLAB optimization toolbox}. 
The function \textit{fmincon} finds the minimum  of a problem specified by
\begin{mini!}
{\substack{$\xv$}}
{L(\xv)}
{}{}
\addConstraint{c(\xv)}{\leq 0}{\quad \text{(nonlinear inequalities)}}
\addConstraint{ceq(\xv)}{= 0}{\quad \text{(nonlinear equations) }}
\label{eq_OptimisationF}
\end{mini!}
$c(\xv)$ and $ceq(\xv)$ are functions that return vectors and $L(\xv)$ is a function that returns a scalar. $L(\xv), ~c(\xv)$ and $ceq(\xv)$ can be nonlinear functions \cite{MathWorks.2020}. For the problem specified in this work, $ceq(\xv)$ contains the vehicle model's momentum equalities (\ref{eq:ImpulseLong},\ref{eq:ImpulseLat}, \ref{eq:ImpulseYaw}) as well as equality constraints (\ref{eq:constraintsYawAcceleration},\ref{eq:constraintsARatio},\ref{eq:constraintsTorqueBraking},\ref{eq:constraintsBrakingBias}).
The inequalities contained in $c(\xv)$ are described by equations (\ref{eq:ineqConstraintsSideSlipAngle},\ref{eq:ineqConstraintsSlipRatio},\ref{eq:ineqConstraintsTorqueMax}).

\subsubsection{\optTire optimisation}
The optTire optimiser is similar to the \optCog optimiser. Instead of the complete vehicle with a state vector \xv, every tire state vector $\yv_\mathrm{as}$ is optimised independently. \autoref{fig:TPERcombined}b visualizes the process. We base the  \optTire optimisation on the \optCog state to make it less dependent on a possibly sub-optimal force distribution from the \Init state (this is especially relevant for the amateur drivers). The \optCog tire state vector is a subset of the \optCog vehicle state vector
\begin{equation}
\yoptCog \subseteq \xoptCog.
\end{equation}

\yoptCog is subdivided into a constant part and a variable part, which is altered by the optimiser
\begin{equation}
\begin{aligned}
 \yoptCog =[ &\yv^\mathrm{oc}_\mathrm{as,const} ~ \yv^\mathrm{oc}_\mathrm{as,var}]^{\textrm{T}},\\
    \yv^\mathrm{oc}_\mathrm{as,const} =[&~Fz_\mathrm{as}~\mu_\mathrm{as}~
    \gamma_\mathrm{as}~r_\mathrm{as}],\\
    \yv^\mathrm{oc}_\mathrm{as,var} =[ &\alpha_\mathrm{as}^\mathrm{oc}~\kappa_\mathrm{as}^\mathrm{oc}].
\end{aligned}
	\label{eq:xInTire}
\end{equation}

The \optTire optimisation results in a new tire state vector
\begin{equation}
\begin{aligned}
	\yv^\mathrm{ot}_\mathrm{as} =[ &\yv^\mathrm{oc}_\mathrm{as,const} ~ \yv^\mathrm{ot}_\mathrm{as,var}]^{\textrm{T}},\\
\yv^\mathrm{ot}_\mathrm{as,var} =[ &\alpha_\mathrm{as}^\mathrm{ot}~\kappa_\mathrm{as}^\mathrm{ot}],\\
	\text{such that } L_{}^{\mathrm{ot}}(\yv^\mathrm{ot}_\mathrm{as} )= & \min L_{}^{\mathrm{ot}}(\yv^\mathrm{}_\mathrm{as} ).
	\end{aligned}
	\label{eq:xOptTire}
\end{equation}
Similar to the vehicle model a tire model \TireMod is used to calculate the relevant variables from the tire state vector \yv, in this case the forces at each tire in x and y direction
\begin{equation}
F_\mathrm{as} = \TireMod_{F_\mathrm{as}} (\yv). 
\label{eq:def_TM}
\end{equation}

The force generated by each tire is maximised individually 
 
\begin{equation}
	L^{\mathrm{ot}}_\mathrm{as}\left(\yv\right)~=~-\sqrt{(F_\mathrm{x,as})^{2}+(F_\mathrm{y,as})^{2}},
	\label{eq:costfunTire}
\end{equation}
while keeping the direction of the force equal to the \optTire state 
\begin{equation} 
	\dfrac{F_\mathrm{x,as}^\mathrm{ot}}{F_\mathrm{y,as}^\mathrm{ot}}-\dfrac{F_\mathrm{x,as}^\mathrm{oc}}{F_\mathrm{y,as}^\mathrm{oc}}~=~0.
\label{eq:constraintsTire}
\end{equation}

Note that there does not have to exist a valid vehicle state for the optimised tire state nor a valid trajectory.  
Similar to the \optCog optimiser, the \textit{MathWorks$^{\textrm{\textregistered}}$ MATLAB optimisation toolbox} with the \textit{Active-Set} solver option is used.

\subsubsection{Scores}
\label{chap_metrics}
As a first step to obtain scores from the \optCog results, the acceleration at the center of gravity and forces at the tires in the xy-plane in a polar coordinate system are calculated
\begin{equation}
\begin{gathered}
a_\mathrm{\rho,cog} = \sqrt{a_\mathrm{x,cog}^2+a_\mathrm{y,cog}^2},\\
a_\mathrm{\phi,cog} = \mathrm{atan}(\frac{a_\mathrm{y,cog}}{a_\mathrm{x,cog}}),\\
F_\mathrm{\rho,as} = \sqrt{F_\mathrm{x,as}^2+F_\mathrm{y,as}^2},\\
F_\mathrm{\phi,as} = \mathrm{atan}(\frac{F_\mathrm{y,as}}{F_\mathrm{x,as}}).\\
\end{gathered}
\end{equation}

Then, the sum of all tire forces is defined as
\begin{equation}
F_\mathrm{\rho,tires} =F_\mathrm{\rho,fl}+F_\mathrm{\rho,fr}+F_\mathrm{\rho,rl}+F_\mathrm{\rho,rr}.
\end{equation}

We define the following metrics which can be interpreted as scores for the system vehicle-driver. 
The ratio between the amount of initial vehicle acceleration at the center of gravity to the \optCog acceleration yields $S_\mathrm{handling}$

\begin{equation}
\begin{gathered}
S_\mathrm{handling} = \frac{a_\mathrm{\rho,cog}^{\mathrm{in}}}{a_\mathrm{\rho,cog}^{\mathrm{oc}}}.\\
\end{gathered}
\end{equation}

The vehicle score is defined as the ratio between the \optCog and \optTire forces
\begin{equation}
\begin{gathered}
S_\mathrm{veh-traj} = \frac{F_\mathrm{\rho,tires}^{\mathrm{oc}}}{F_\mathrm{\rho,tires}^{\mathrm{ot}}}.\\
\end{gathered}
\end{equation}

Finally, the overall score emerges from the \Init forces to the \optTire forces
\begin{equation}
\begin{gathered}
S_\mathrm{tot} = \frac{F_\mathrm{\rho,tires}^{\mathrm{in}}}{F_\mathrm{\rho,tires}^{\mathrm{ot}}}.\\
\end{gathered}
\end{equation}





\subsubsection{Limitations}

Since we apply the optimisation method on a relatively large data set and aim to use the methods for race preparation in future, computation time is a concern.  This is also a main reason why we introduce the constraint of a constant driving line which is a simplification since the driver's performance on the stabilisation level also influences the driving line. For example when entering a corner too fast, the vehicle might get pushed towards the outside, although not being intended by the driver. However, for the professional drivers we can assume that they anticipate this effect. This is supported by \citet{Macadam.2003} who stated that the drivers choice of action is not only based on the current state but also prior knowledge and experience.\\
We execute the optimisation for approximately $3.5$ million data instances. The calculation time for one lap with an average length of $80$ seconds is approximately $200$ seconds on computer with two 10-core \emph{Intel\textregistered Xeon\textregistered Silver 4144} CPUs. 
To reach this speed, both optimiser rely on parallel processing which significantly reduces calculation time. As a consequence, the models compute each data point isolated from its temporal surrounding. Dynamic effects due to changed accelerations from the optimised states are not respected, for example a change in load transfer. However, the closer the \Init state is to the dynamic limit (the \optCog state) the smaller the error will be, which is where race drivers usually operate the vehicle. Though, for some applications a calculation time of $200$ seconds per lap is still too high. For example in race simulations 50 and more laps can be driven and a very quick analysis is needed. The resulting calculation time of $10000$ seconds would make it infeasible in practice. Ideal would be a real-time calculation. Therefore, we propose an end-to-end machine learning approach to replace the computational expensive optimisation. 

\subsection{Machine learning predictor}
We propose a machine learning model as an approximation to the aforementioned optimisation approach. Depending on the computer hardware and model size the runtime speed improvement is more than one order of magnitude and the model is real-time application suited. The three scores defined in section \ref{chap_metrics} are predicted simultaneously by an artificial neural network \PredMod based on $\dot{\bm{x}}^\mathrm{p}$, which is a part of the initial vehicle state vector \xInit 
\begin{equation*}
[\tilde{S}_\mathrm{handling}~\tilde{S}_\mathrm{veh-traj}~\tilde{S}_\mathrm{tot}] = \PredMod (\dot{\bm{x}}^\mathrm{p}).
\end{equation*}


\subsubsection{Artificial neural network architecture}

\begin{figure}
\includegraphics[width=0.9\textwidth]{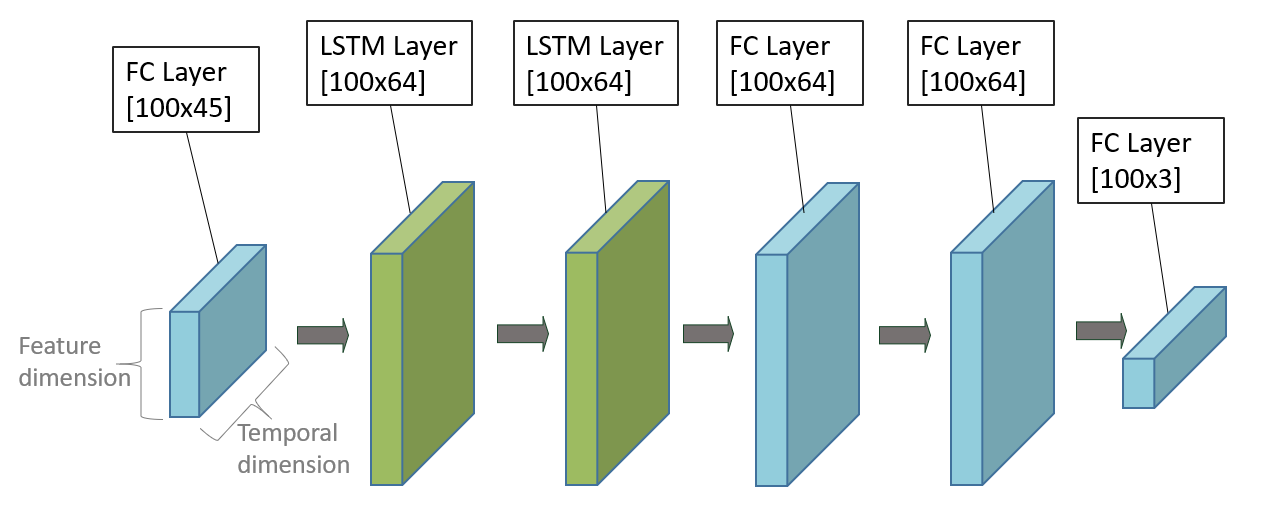}
\caption{Machine learning predictor model architecture. Two LSTM and two fully connected layers constitute the hidden layers. }
\label{fig_lstm}
\end{figure}

The model is based on a LSTM neural network architecture, which has been used for time series prediction on a motorsport data set before \cite{Schleinitz_vasp.2021}. The hyperparameters are listed in \autoref{tab_lstm_params} and the architecture in \autoref{fig_lstm}. The model is built in R 3.6.2 \cite{RDevelopmentCoreTeam.2008} using the R interface to keras \cite{JJAllaire.2018} and tensorflow \cite{JJAllaire.2018b}.  \\
 The LSTM was introduced by \citet{Hochreiter.1997}. A LSTM consists of an \emph{input gate}, \emph{forget gate}   and \emph{output gate}. With these, the LSTM can decide which information will be forgotten, learned or passed on to the next time step. The outputs of the previous time step  serve as inputs for the current time step. Apart from the output, a LSTM passes a cell state to the next time step. The cell state makes it easier for information to flow unchanged through multiple time steps. This results in the ability of the LSTM cells to learn long-term dependencies while avoiding the vanishing /exploding gradient problem \cite{vanHoudt.2020, Schleinitz_vasp.2021}.

\begin{table}[b]
	\centering 
	\caption{Hyperparameters for the LSTM-based predictor module}
	\small
	\begin{tabular}{l l}
		 \hline
        \textbf{Parameter}     & \textbf{Setting}         \\  
        \hline
        Learning rate         & 0.001             \\ \hline
        Batch size         & 128               \\ \hline
         Dropout         & 0.3               \\ \hline
          Recurrent dropout       & 0.1               \\ \hline
          Optimiser         & "Adam" \cite{Kingma.22.12.2014}               \\\hline
          Input temporal dimension  & 100\\ \hline
           Input feature dimension  & 24\\ \hline
           Output temporal dimension  & 100\\ \hline
           Output feature dimension  & 3\\ 
        \hline

	\end{tabular}
	\label{tab_lstm_params}
\end{table}

\subsubsection{Data preparation}
All time series data are normalized to increase the numerical stability. For this purpose let $\bm{w} \in \mathbb{R}^{m}$ be a raw time series signal. Then, $\bm{x} \in \mathbb{R}^{m}$ is calculated by
\begin{equation*}
\bm{x}= \frac{1}{\sigma(\bm{w})}(\bm{w}- \mu(\bm{w})\bm{e} ),
\end{equation*}
with  the all-ones vector $\bm{e} \in \mathbb{R}^{m}$, whereby $\mu(\bm{w}) \in \mathbb{R}$ is the sample mean and $\sigma(\bm{w}) \in \mathbb{R}$ the sample standard deviation of $\bm{w}$.  

The data is split in  training ($80\%$), validation ($10\%$) and test set ($10\%$). Also, the Track T2 is excluded from the training set to examine the model accuracy on an unseen track.

\subsubsection{Model selection}
\label{chap_model_sel}
The root-mean square error (RMSE) between  prediction and  reference is used as an evaluation criteria for the neural network. The RMSE between a reference signal $\bm{yr}$ and a predicted signal $\bm{y}$ with $m$ time steps is
\begin{equation*}
\text{RMSE}(\bm{yr},\bm{y})=\sqrt{\frac{1}{m} \sum\limits_{t=0}^{m} (\bm{yr}^{(t)}-\bm{y}^{(t)})^2 }.
\end{equation*}

Three different models M1, M2 and M3 are trained on the data set. These models have different input time series. All inputs are obtained from the initial vehicle state.

\begin{itemize}
\item M1 (32 inputs): $\alpha_{\mathrm{as}}^{\mathrm{}}$, $\kappa_{\mathrm{as}}^{\mathrm{}}$, $Fx_{\mathrm{as}}^{\mathrm{}}$, $Fy_{\mathrm{as}}^{\mathrm{}}$, $Fz_{\mathrm{as}}^{\mathrm{}}$, 
$\pBrake$, $\rThrottle$, $\dot{\psi}$, $a_\mathrm{x,cog}$, $a_\mathrm{y,cog}$, $\delta$, $v$, $\beta$, $\aRocker$
\item M2 (45 inputs): $\alpha_{\mathrm{as}}^{\mathrm{}}$, $\kappa_{\mathrm{as}}^{\mathrm{}}$, $Fx_{\mathrm{as}}^{\mathrm{}}$, $Fy_{\mathrm{as}}^{\mathrm{}}$, $Fz_{\mathrm{as}}^{\mathrm{}}$, 
$\pBrake$, $\rThrottle$, $\dot{\psi}$, $a_\mathrm{x,cog}$, $a_\mathrm{y,cog}$, $\delta$, $v$, $\beta$, $\aRocker$, $\gamma_\mathrm{as}$, $r_\mathrm{as}$, $\mu_\mathrm{as}$
\item M3 (16 inputs): $Fz_{\mathrm{as}}^{\mathrm{}}$, 
$\pBrake$, $\rThrottle$, $\dot{\psi}$, $a_\mathrm{x,cog}$, $a_\mathrm{y,cog}$, $\delta$, $v$, $\beta$, $\aRocker$
\end{itemize}

\begin{figure}
\begin{subfigure}[c]{0.5\textwidth}
\includegraphics[width=1\textwidth]{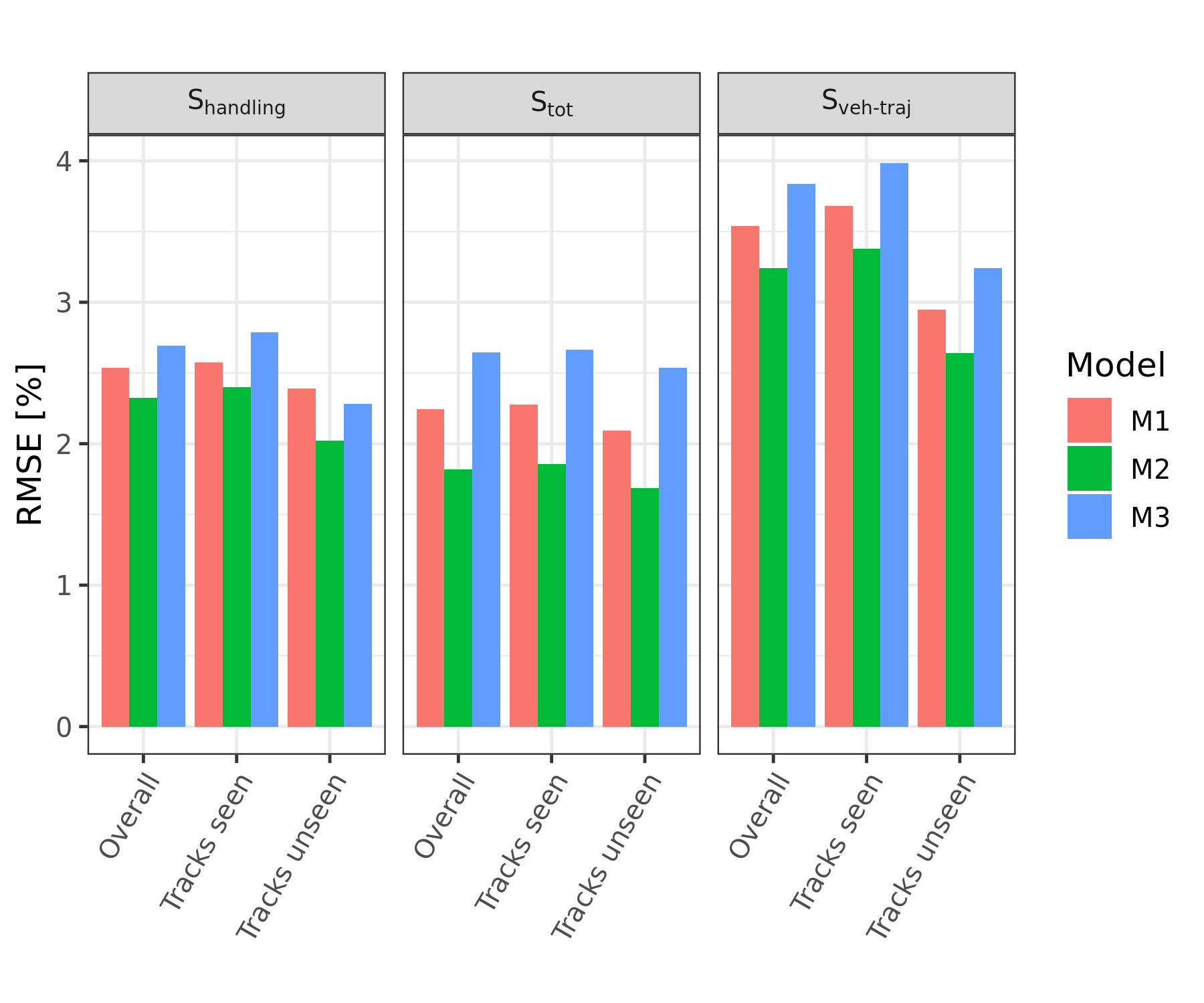}
\subcaption{Prediction model accuracy over tracks. The tracks are split into seen tracks, which were part of the training set and unseen tracks which were not. There is no significant RMSE difference between the two cases. }
\end{subfigure}
\begin{subfigure}[c]{0.5\textwidth}
\includegraphics[width=1\textwidth]{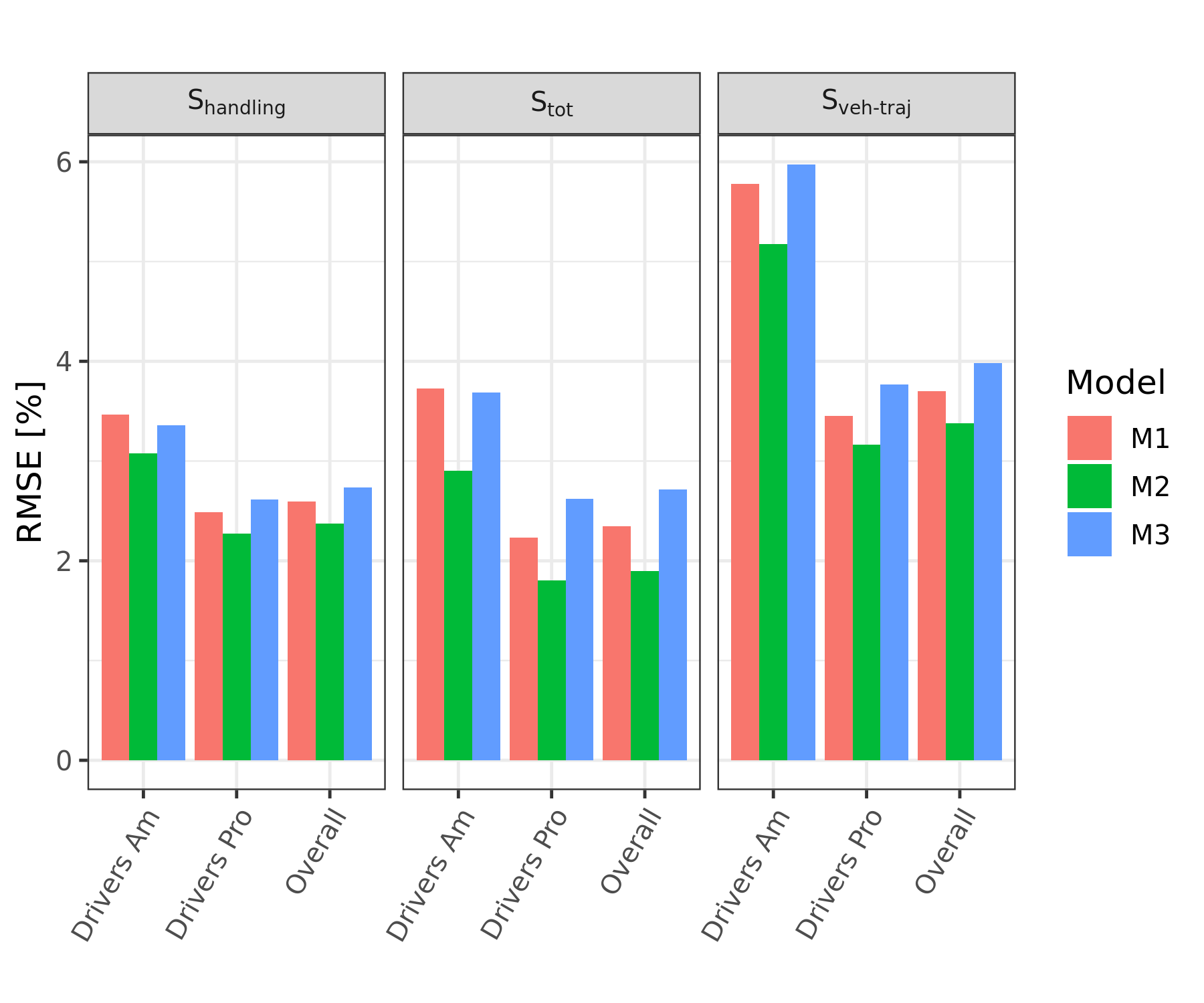}
\subcaption{Model accuracy over drivers. The professional drivers 'Drivers Pro' were part of the training set, whereas the amateur drivers 'Drivers Am' were not. The error for the amateur drivers is clearly higher, however, still in the same range. }
\end{subfigure}
\caption{In all cases the model M2 has the lowest error. The error for the prediction of \SVeh is generally higher than  for \SHand or \STot. }
\label{fig_lstm_res}
\end{figure}

The models' accuracy should be consistent for different factors of influence. \autoref{fig_lstm_res} shows the RMSE in dependence of tracks as well drivers.  To detect possible overfitting to the training data, we compare the RMSE on seen and unseen tracks (which were not a part of the training set) in \autoref{fig_lstm_res} a. None of the models showed overfitting to the seen tracks, the error on the unseen track is even a bit lower. To further test the models, they are also tested on the amateur drivers' data. The models were only trained on data from the professional drivers, which have distinctly higher scores. The amateur driver scores are therefore well outside the scope of the training data. \autoref{fig_lstm_res} depicts the resulting RMSE. The error is as expected higher  but  the models  are very usable for the task at hand. This indicates that the models are robust. 
In all cases the model M2 has the lowest error and is therefore chosen for this work. The additional inputs compared to M1 and M3 contained additional information which helped the model to predict more accurately. Since the advantage of M2 is also visible for unseen tracks and the amateur drivers it is not a result of overfitting.


\subsection{Control states}
In order to analyse  driver performance  independent from a specific track section, control states are defined based on  the driver's inputs. Three boolean variables describe if  the brake pedal, throttle pedal or steering wheel angle $\delta$ are active
\begin{equation}
B_{brake}= 
\begin{cases}
   1 ,& \text{if } \pBrake>10 bar\\
    0, & \text{otherwise,}
\end{cases}
\end{equation}

\begin{equation}
B_{throttle}= 
\begin{cases}
   1 ,& \text{if } \rThrottle>10 \%\\
    0, & \text{otherwise,}
\end{cases}
\end{equation}

\begin{equation}
B_{steer}= 
\begin{cases}
   1 ,& \text{if } \vert \delta \vert>10^\circ \\
   1, & \text{if } \vert\dot{\delta}\vert > 500^\circ/s\\
    0, & \text{otherwise.}
\end{cases}
\end{equation}

%
%

The second condition for $B_{steer}$ ensures that occasions  where the steering angle is small but the change rate is high, are included. This can be for example during a quick counter steer. \\
The combination of the boolean variables results in the four control states. These relate to the commonly defined cornering sections \cite{goy.16}, which are in parentheses:
\begin{itemize}
\item Pure brake (Braking): $B_{brake}=1, ~ B_{throttle}=0, ~ B_{steer}= 0$ 
\item Trail brake (Turn Entry): $B_{brake}=1, ~ B_{throttle}=0, ~ B_{steer}= 1$ 
\item Pure steer (Mid Corner): $B_{brake}=0, ~ B_{throttle}=0, ~ B_{steer}= 1$ 
\item Throttle steer (Turn Exit) : $B_{brake}=0, ~ B_{throttle}=1, ~ B_{steer}= 1$ 
\end{itemize}

The other possible combinations are not relevant for the driver analysis in the context of this paper. For example under pure throttle the vehicle's acceleration  is usually  limited by the engine and not the driver. An overview of the control states in the time domain is e.g. depicted in \autoref{fig_kappa}.

\section{Results}

\subsection{\optTire optimisation}
Comparing the \Init state to the \optTire state leads to insights how the vehicle is handled at the dynamic limit. We look at the initial and optimised slip ratios and slip angles of the tires. 

\subsubsection{Slip ratio $\kappa$}
\label{chap_kappa}
\begin{figure}
\begin{subfigure}[c]{1\textwidth}
\begin{center}
\includegraphics[width=0.7\textwidth]{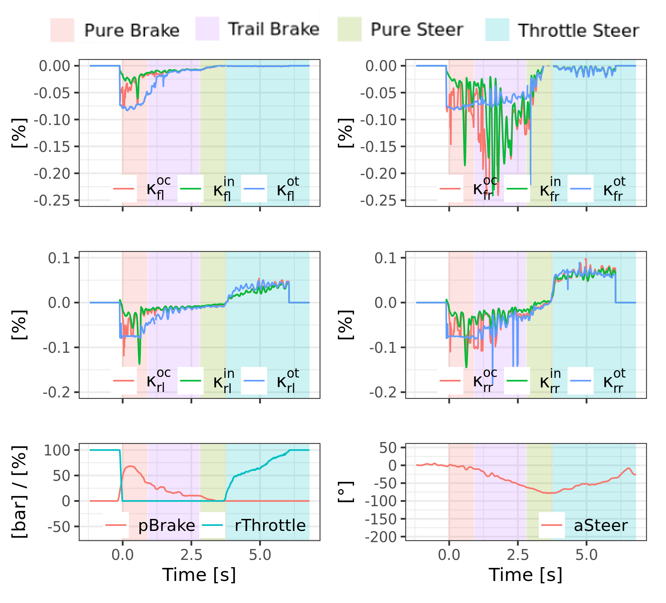}
\end{center}
\subcaption{Exemplary overview about $\kappa$ in the time domain for the \Init, \optCog and \optTire state.}
\end{subfigure}
\begin{subfigure}[c]{1\textwidth}
\begin{center}
\includegraphics[width=0.7\textwidth]{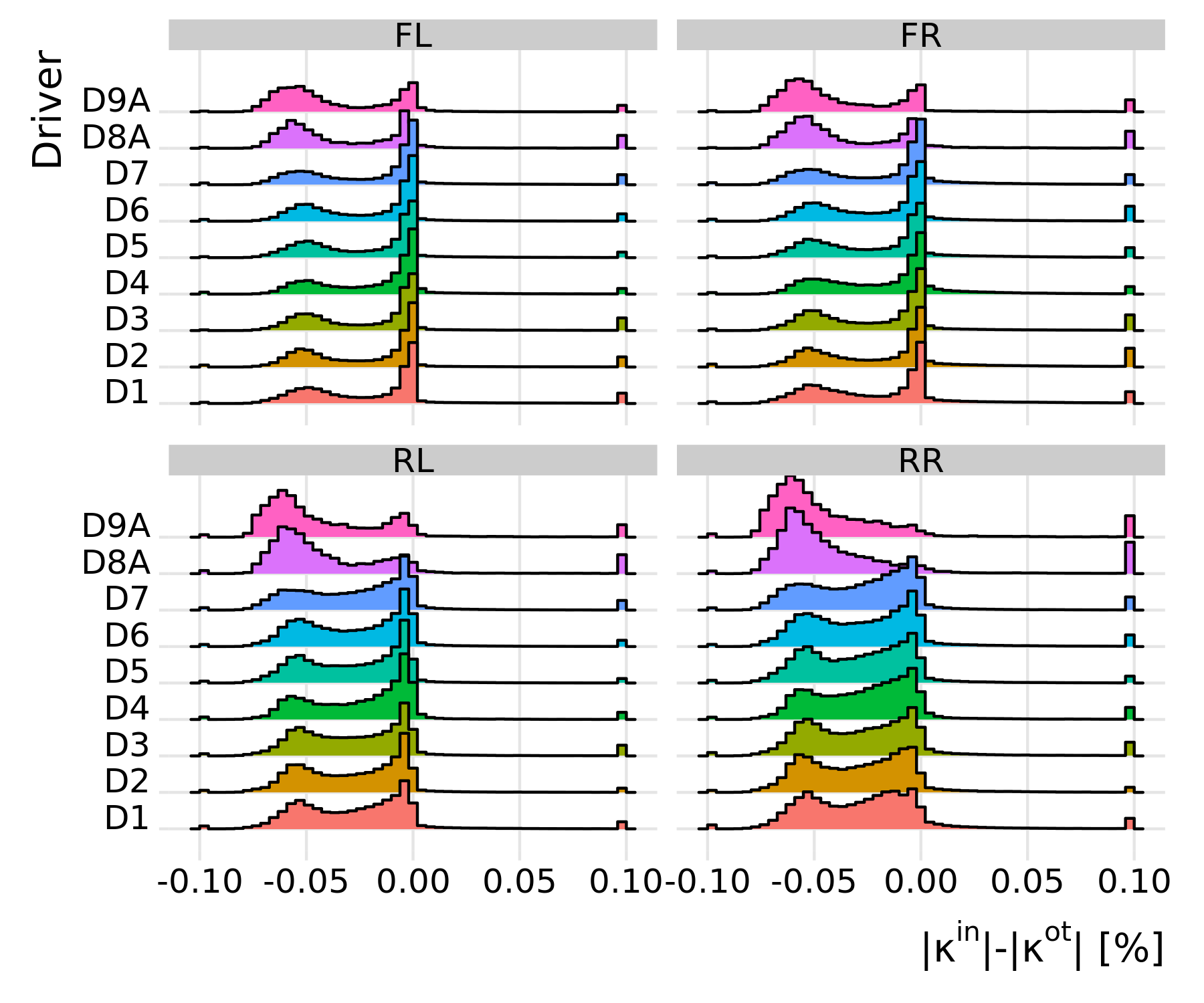}
\end{center}
\subcaption{ $\kappa$ difference distribution between the \Init and \optTire state for the complete data set over the four tires and drivers. Negative values indicate that the absolute of $\kappa^\mathrm{in}$  is smaller than the absolute of $\kappa^\mathrm{ot}$.}
\end{subfigure}
\caption{The histograms (bottom plot) show the drivers very rarely exceed the optimal Slip Ratio $\kappa$. The top plot shows such rare occasion for the front right wheel. }
\label{fig_kappa}
\end{figure}

\autoref{fig_kappa}a shows that the slip ratio in the  \optCog state $\kappa^\mathrm{oc}$ changes with a higher frequency in the Pure Brake zone than the \Init state $\kappa^\mathrm{in}$. The \optTire state $\kappa^\mathrm{ot}$ which depicts the optimum slip ratio for each tire provides a reference. This optimum cannot be reached for all tires simultaneously, therefore the \optCog optimiser has to make trade-offs by alternating the tires that reach the optimum.  
\autoref{fig_kappa}b shows the distribution of slip ratio differences from the \Init to the \optTire state per tire. It seems that all drivers are very careful not to exceed the optimal value for $\kappa$, since that would lead quickly to a so called 'tire lock-up' in reality.  That means the tire is no longer rotating while the vehicle is still moving and just rubbing over the road surface. A tire lock-up not only decreases the braking acceleration but also causes significant tire wear due to the high relative velocities, which in turn can significantly decrease performance for the following laps. For comparison the amateur drivers are shown as well.  The center of the distributions is further away from the optimum both on the front and on the rear axle. We conclude that the amateur drivers are more cautious  and cannot reach the optimum as well as the the professional drivers. 

\subsubsection{Slip angle $\alpha$}
\begin{figure}
\begin{subfigure}[c]{1\textwidth}
\begin{center}
\includegraphics[width=0.7\textwidth]{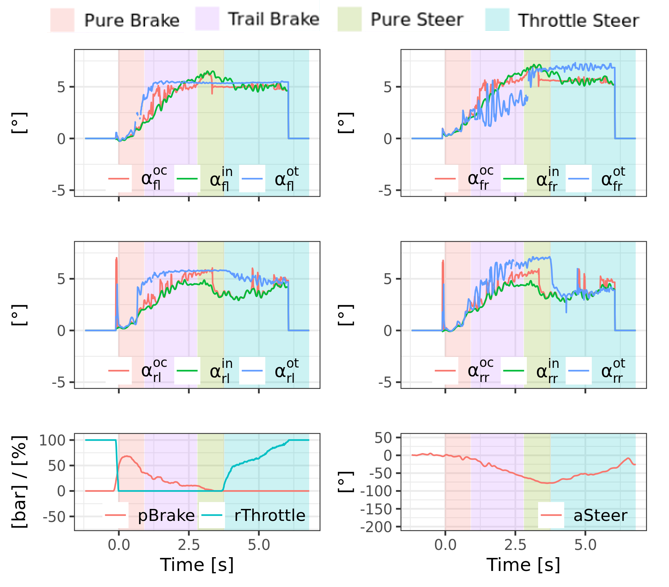}
\end{center}
\subcaption{Exemplary overview about $\alpha$ in the time domain for the \Init, \optCog and \optTire state.}
\end{subfigure}
\begin{subfigure}[c]{1\textwidth}
\begin{center}
\includegraphics[width=0.7\textwidth]{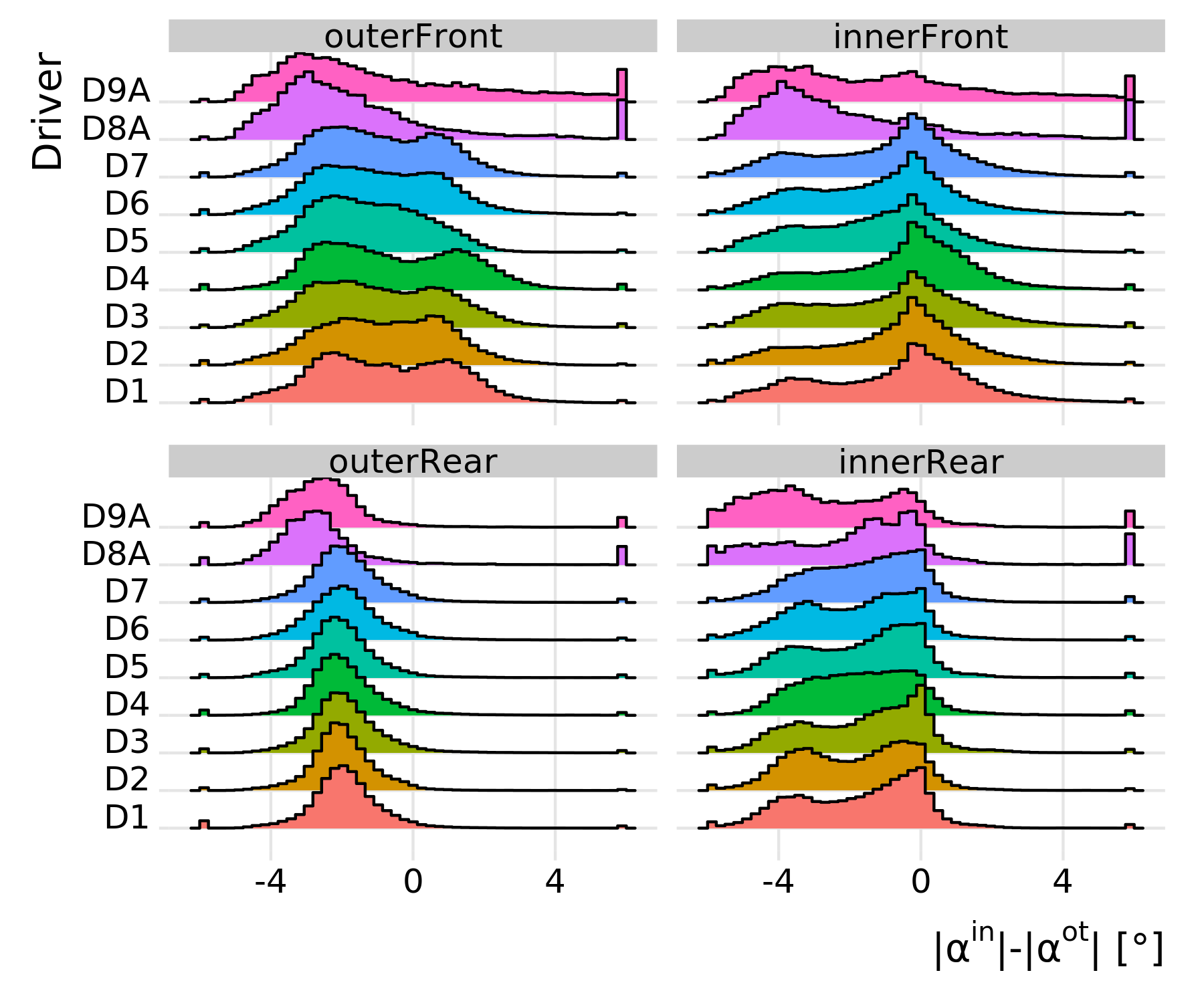}
\end{center}
\subcaption{Distribution of slip angle differences between the \Init and \optTire state over the tires and drivers for all sections where $|gLat| \geq 1.5g$.  Negative values indicate that the absolute of $\alpha^\mathrm{in}$  is smaller than the absolute of $\alpha^\mathrm{ot}$. The panel is split into outer and inner tire instead of left and right tire to make it track independent.} 
\end{subfigure}
\caption{Both panels show that the drivers often exceed the optimal slip angle $\alpha$ on the front axle, while usually staying below the limit at the rear axle. Especially on the outer rear axle, the drivers avoid overshooting the optimum. Differences between the professional and amateur drivers are clearly visible. }
\label{fig_alpha}
\end{figure}

\autoref{fig_alpha} gives a similar overview for the slip angle $\alpha$. \autoref{fig_alpha}a shows an example in the time domain. For the front axle (the two topmost plots) the \Init and \optCog state are close to each other, on the rear axle seems to be a larger gap.  Over the complete data set the histograms in \autoref{fig_alpha}b draw a similar picture. Instead of left and right the panel is split into inner and outer tire to make the analysis independent of a specific race track. What is more, during a cornering manoeuvre the outer tires have a higher normal force due to load transfer. \\
All professional drivers exceed the optimal slip angle $\alpha^\mathrm{ot}$ at the front axle most of the time. For the outer front the distribution is bimodal for most drivers with a maximum at approx. $-2^\circ$ and $1.5^\circ$. Interestingly, on the inner front there is a more pronounced maximum a $0^\circ$. On the outer rear all drivers clearly avoid slip angles higher than the optimum, resulting in a peak at $-2^\circ$. The distribution on the inner rear has a wider range and reaches slightly in the positive values. 
 The explanation for not exceeding the optimum on the outer rear axle is stability. Loosing grip on the rear axle would lead to an unstable oversteer and since the outer rear provides most of the grip it is more important to not exceed the optimum there than on the inner rear. On the front however, the drivers intentionally drive at higher slip angles than would be optimal because the generated lateral force stays relatively constant from the optimum towards higher slip angles but has a steeper slope from small values to the optimum. To better control the vehicle and anticipating it's reaction, the drivers stay in the more constant region above the optimum on purpose. \\
Between the professional drivers, some differences can be observed. On the outer and inner front tire, Driver D5 has a more unimodal distribution than the rest. On the inner rear tire, the same holds for Driver D4. The amateur drivers which are included as a reference have a distinctly different distribution. On the front, they have smaller slip angles than the optimum in general, however, instances where they have much higher slip angles than the optimum are also more frequent than for the professional drivers. On the rear axle the situation is similar. Analogical to the slip ratio, the amateur drivers are more cautious not to exceed the optimal slip angles. \\

\subsection{\optCog optimisation}
\begin{figure}
\begin{center}
\begin{subfigure}[c]{0.8\textwidth}
\begin{center}
\includegraphics[width=0.8\textwidth]{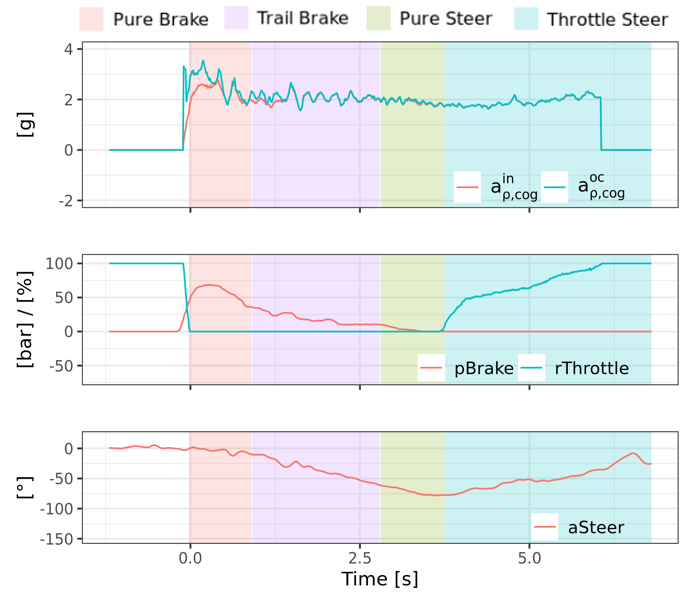}
\end{center}
\subcaption{Exemplary overview about the accelerations from the \Init and \optCog state in the time domain as well as the brake, throttle and steering inputs from the driver. The background is coloured with respect to the defined control states.}
\end{subfigure}
\begin{subfigure}[c]{0.8\textwidth}
\begin{center}
\includegraphics[width=0.9\textwidth]{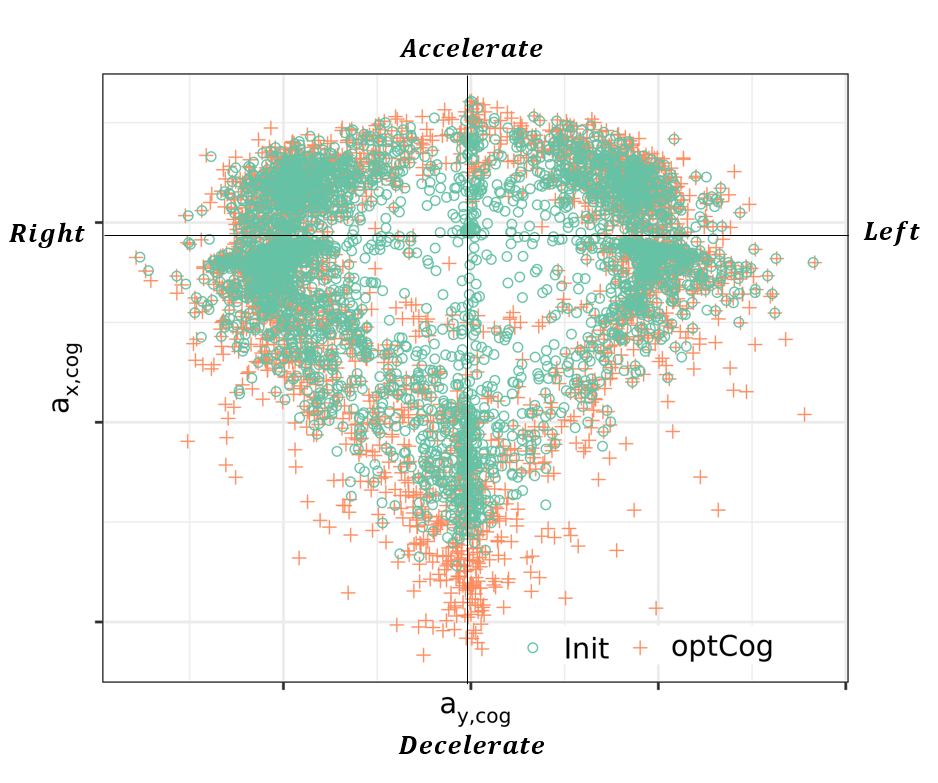}
\end{center}
\subcaption{'GG-Diagram' of the accelerations in the xy-plane for the \Init and \optCog state. Under pure braking (decelerate) there is a significant difference between the two states.}
\end{subfigure}
\end{center}
\caption{\optCog versus \Init state accelerations.}
\label{fig_acc_example}
\end{figure}

\autoref{fig_acc_example}a shows an exemplary overview of the acceleration from the \optCog optimiser. It is close to the \Init acceleration in the phases Pure Steer and Throttle Steer. For Trail Brake, the difference is slightly larger and for Pure Brake the \optCog acceleration is significantly higher. The different control states are indicated by the coloured background. \\
A common visualization method in motorsport for accelerations is the 'GG-Diagram'; \autoref{fig_acc_example}b depicts the acceleration in x- and y-direction.  It allows an understanding of the physical limits of the vehicle \cite{Kegelman.2018}, assuming it was driven on the dynamic limit.  The shape results from the frictional properties of the tires and their kinematic relations to each other (aerodynamic effects like downforce play a role as well). The symmetry of the ellipse is broken for acceleration and deceleration since the acceleration of the vehicle is not always limited by tire grip but also the engine power. In contrast, the brakes are designed to exert enough torque to use the full friction potential. \\
The comparison between the \Init and \optCog state shows large differences under Pure Brake (Decelerate) and slight differences under Trail Brake (between Decelerate and Right or Left). The other areas match closely.\\
We expect two factors to be responsible for the unrealistic high braking accelerations of the \optCog state. First, the optimiser could directly control the tire slip ratio $\kappa$ with the stated torque constraints. Since the optimiser has no information in the temporal domain, it leads to the high frequency changes in $\kappa$  which could not be achieved in reality. Brake system compliance is also not considered due to this approach. Second, the drivers leave a margin to the optimal tire slip ratio to avoid lock-ups as discussed in Section \ref{chap_kappa}. 

\subsection{Scores evaluation}
\begin{figure}
\begin{center}
\begin{subfigure}[c]{0.8\textwidth}
\begin{center}
\includegraphics[width=0.96\textwidth]{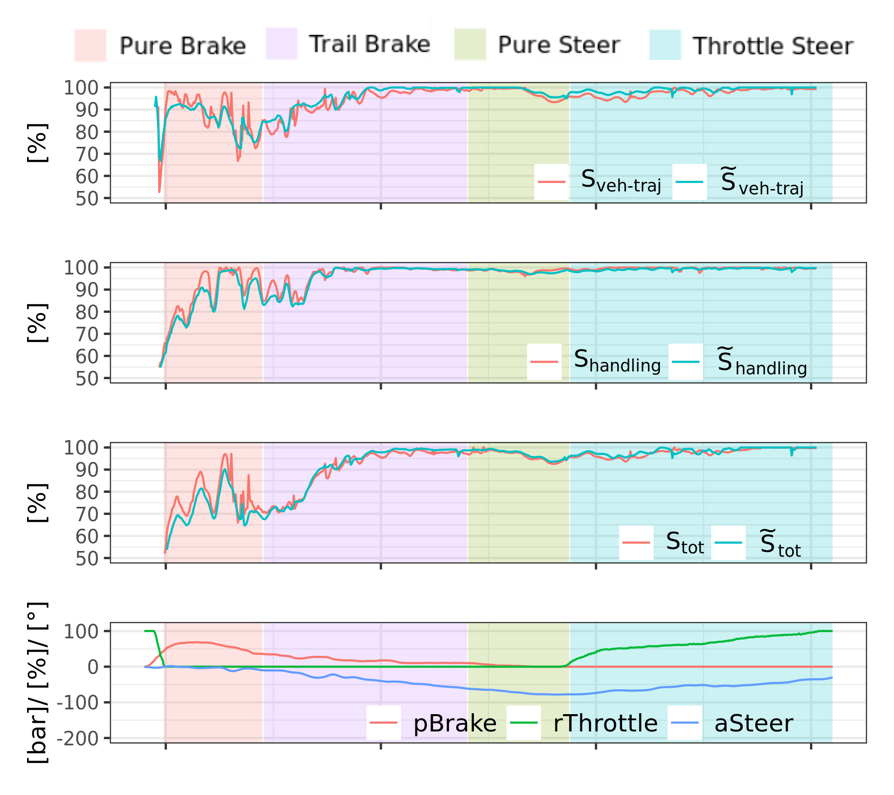}
\end{center}
\subcaption{Exemplary overview about the defined scores and the machine learning predictions in the time domain.}
\end{subfigure}
\begin{subfigure}[c]{0.8\textwidth}
\begin{center}
\includegraphics[width=0.96\textwidth]{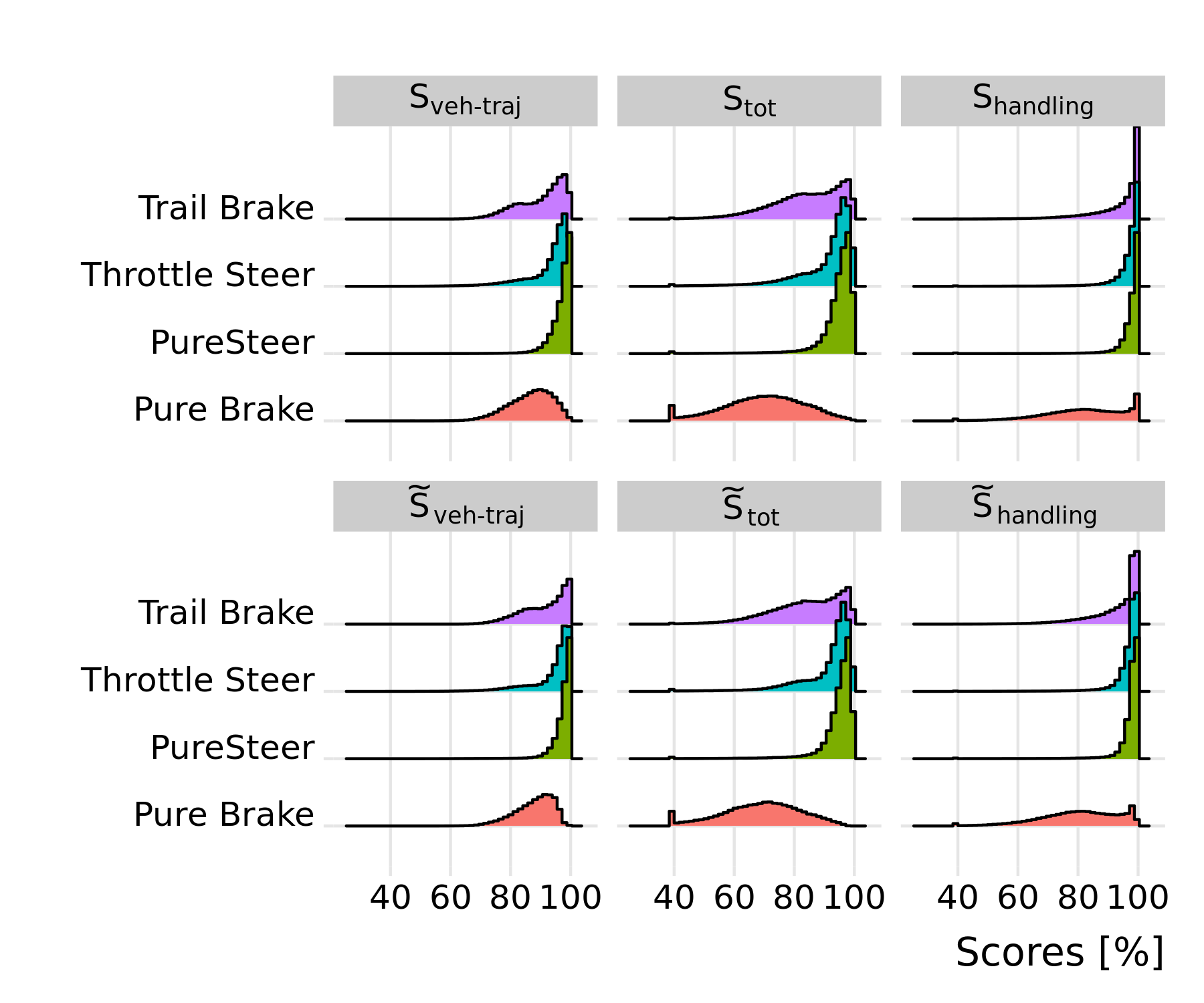}
\end{center}
\subcaption{Scores overview for the complete data set over the control states.  Scores smaller than $40\%$ are grouped in the leftmost bin. Pure Brake and Trail Brake have a higher density in the lower scores. }
\end{subfigure}
\end{center}
\caption{Under braking the distribution of all scores and especially \SHand is skewed towards smaller values.}
\label{fig_scores_example}
\end{figure}

 \autoref{fig_scores_example} shows an overview of the scores and the corresponding predictions in the time domain as well the distributions for the complete data set. As analysed previously, the ratio between the \Init and the \optCog acceleration \SHand is smaller for Pure Brake and Trail brake. 
 
\autoref{fig_scores_example} b shows the differences given the control state. \SHand is spread much wider under Pure Brake than for the other three control states. What is more, the center of the distribution is shifted towards the left. For $S_\mathrm{tot}$ the situation is similar only that for Pure Brake $100\%$ are almost never reached and that the Trail Brake score is skewed to the left. 

This difference makes the control states, which were originally only defined to simplify the analysis for the drivers, necessary to compare different track sectors to each other. For example a track with more Pure Brake sections would lead to lower overall scores in comparison to a track with fewer of such sections. However, by analysing the scores for each control state individually, this will be compensated, resulting in a track independent benchmark. This is a valuable tool for engineers and race drivers which would not be possible without the presented methods.

\subsection{Driving style evaluations}
We provide an exemplary evaluation of the motorsport dataset to show how the presented methods are used and to give an insight into the differences between professional race drivers when controlling a vehicle at its dynamic limit.

\subsubsection{Traditional lap-based metrics}

%

\begin{figure}
\begin{center}
\begin{subfigure}[c]{0.8\textwidth}
\begin{center}
\includegraphics[width=0.88\textwidth]{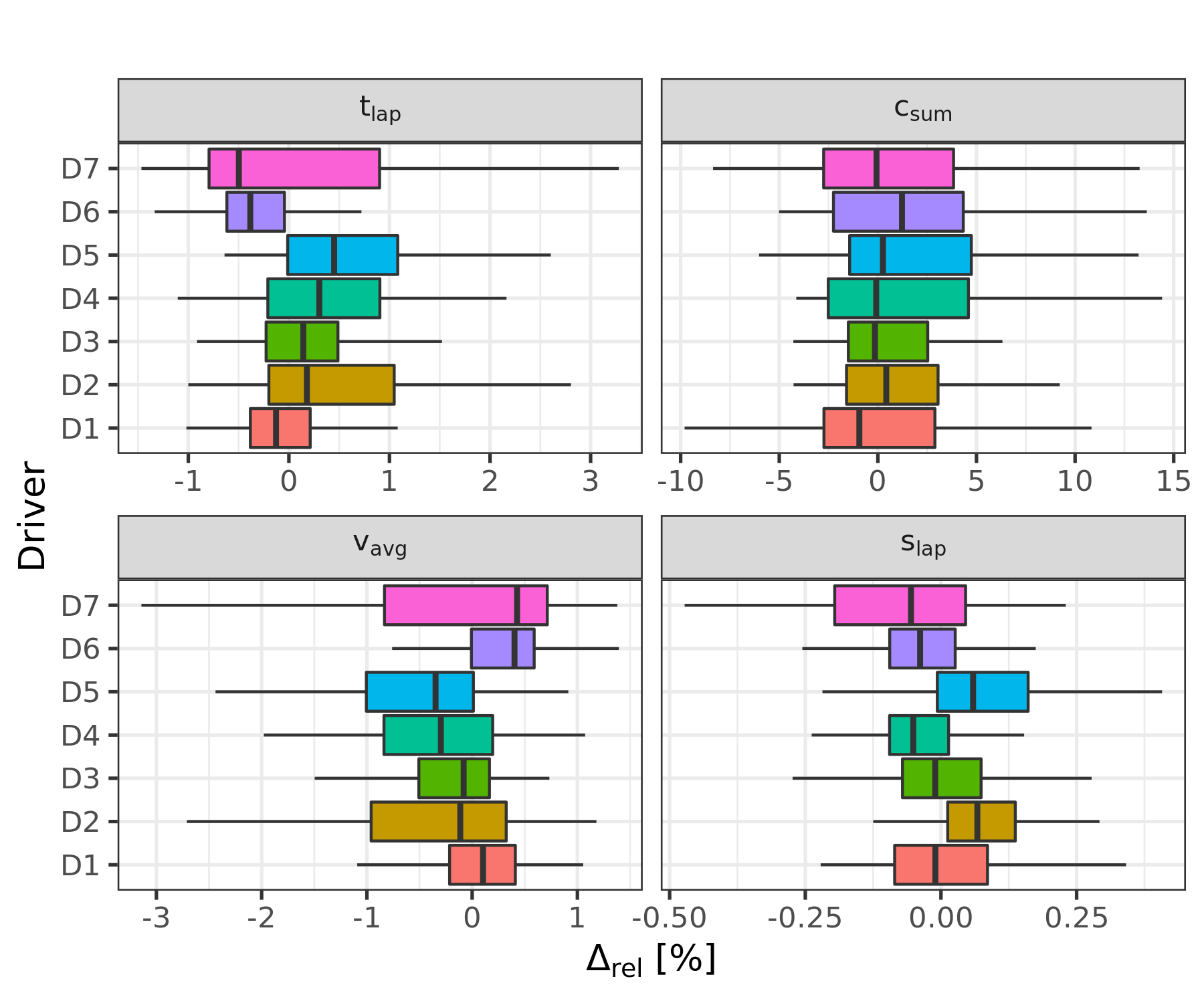}
\end{center}
\subcaption{Professional drivers}
\end{subfigure}
\begin{subfigure}[c]{0.8\textwidth}
\begin{center}
\includegraphics[width=0.88\textwidth]{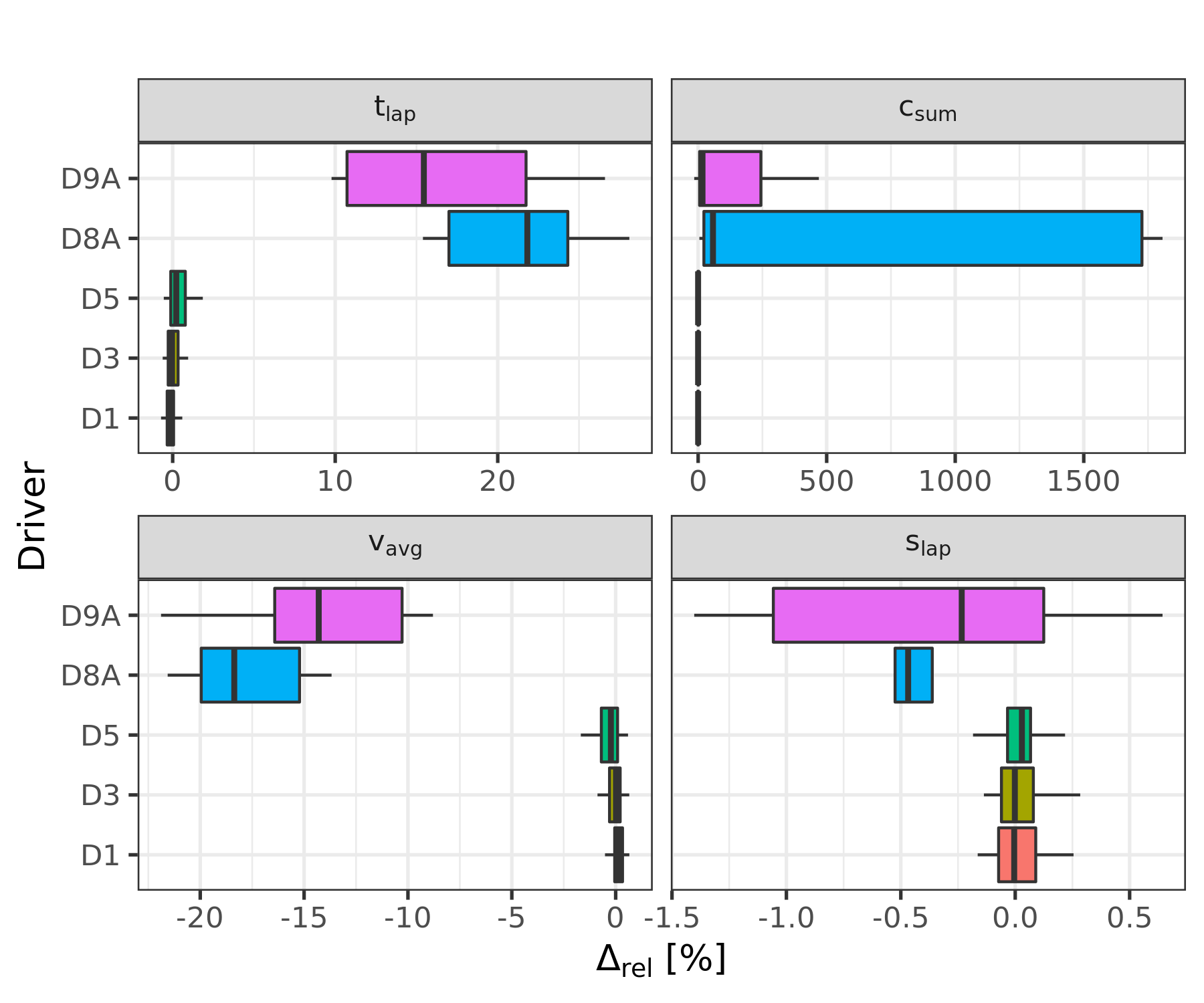}
\end{center}
\subcaption{Amateur versus professional drivers}
\end{subfigure}
\end{center}
\caption{Important metrics on a per-lap basis. All metrics are normed, the units are deviations from the median in $\%$. $tLap$ is the lap time, $v_\mathrm{avg}$ the average speed, $c_\mathrm{sum}$ the sum of the trajectory's curvature and $sLap$ the lap distance. There are significant differences between professional and amateur drivers.}
\label{fig_track_metrics}
\end{figure}

It showed that the most useful representation are boxplots depicting certain characteristics from the distribution of scores. Of interest are the first quartile, the median  and the third quartile, in other words the $25,~50$ and $75\%$ quantile. At first, traditional metrics are compared in \autoref{fig_track_metrics}a on a per-lap basis. The lap time $t_\mathrm{lap}$ is probably the most important criteria to evaluate performance, however it provides no insights on how it is achieved. Closely related are the average velocity $v_\mathrm{avg}$ and the lap distance $s_{lap}$. Additionally, the integral over the absolute curvature $c_\mathrm{sum}$ which is obtained from the trajectory, is shown. 

Looking at the medians, Driver D7 achieved the best laptimes, followed by driver D6. However, the variance is much higher for D7 according to the high spread of the box and the whiskers. This example shows already that the question 'which driver is the best' cannot be answered in general but depends on the circumstances. Differences are also present in $s_\mathrm{lap}$. Driver D2 and D5 chose overall a longer driving line than the rest. The average speed results from lap time and lap distance naturally. 

Although powerful, the disadvantage of these set of metrics is twofold. First, the most important metric $t_\mathrm{lap}$ can be used on a lap basis only. It is possible to use sector times by dividing the tracks in multiple parts, however, there is a limit on how short they can reasonably be. Second, these metrics do not always show where and why lap time differences occur.

\subsubsection{Professional race drivers}
\begin{figure}
\begin{center}
\begin{subfigure}[c]{0.8\textwidth}
\begin{center}
\includegraphics[width=0.88\textwidth]{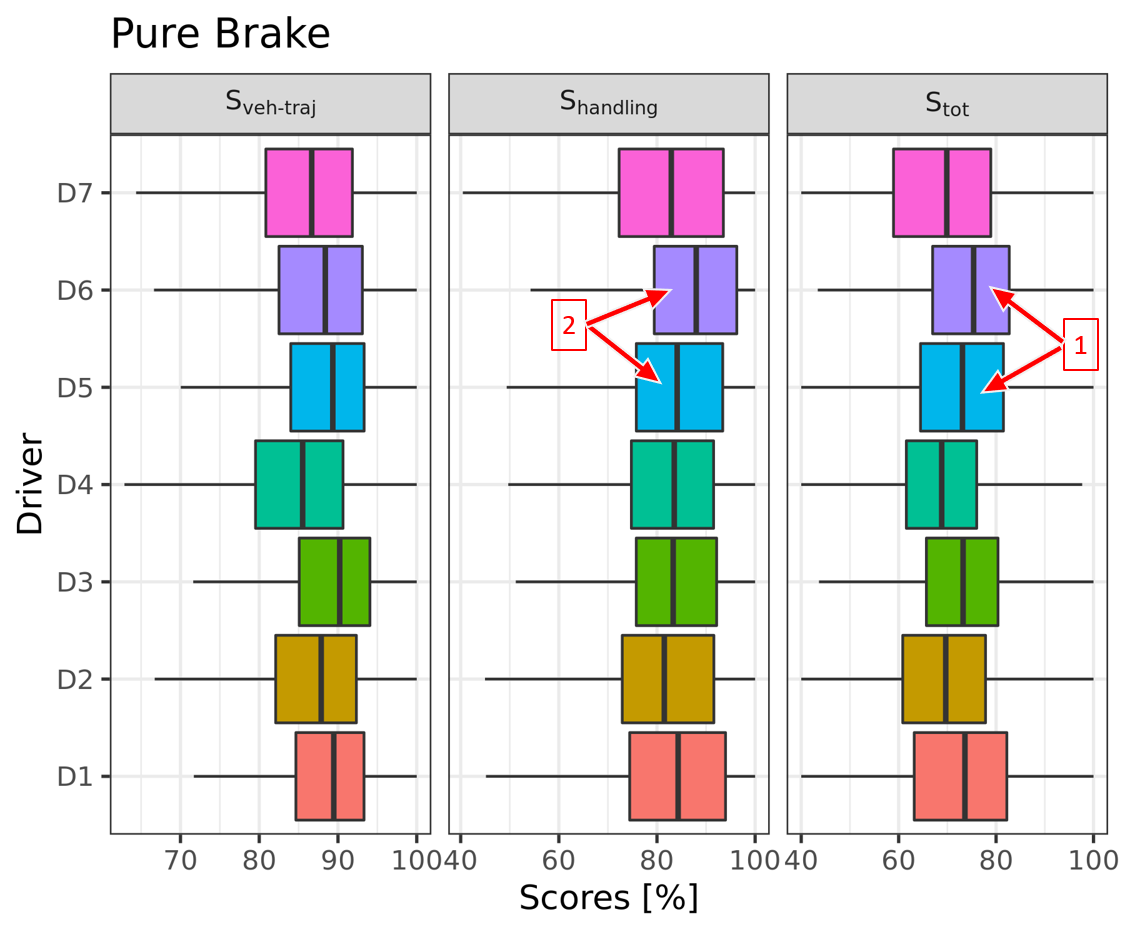}
\end{center}
\end{subfigure}
\begin{subfigure}[c]{0.8\textwidth}
\begin{center}
\includegraphics[width=0.88\textwidth]{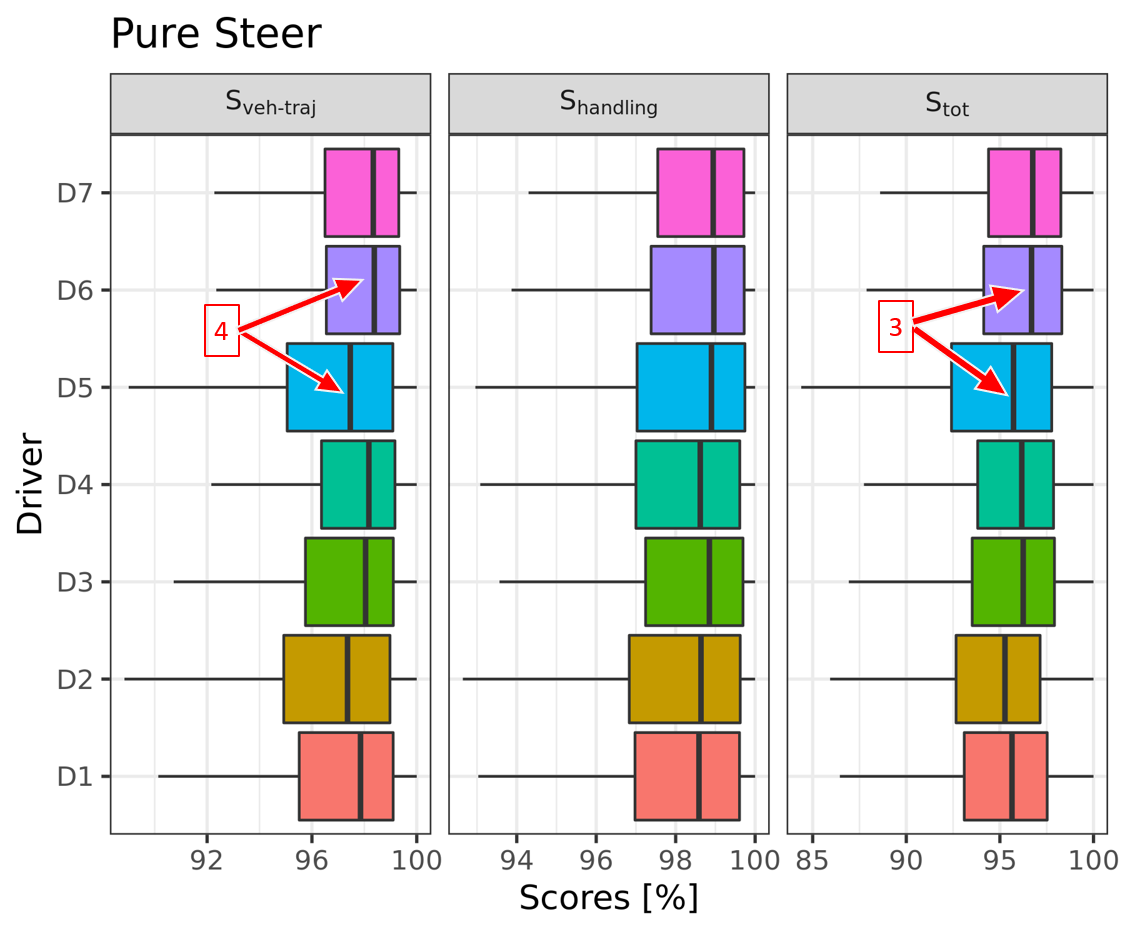}
\end{center}
\end{subfigure}
\end{center}
\caption{Exemplary score overview for the pure brake and pure steer states. The remarks show differences between the drivers D5 and D6.}
\label{fig_overall:scores1}
\end{figure}

We show a visualization that can be used for a single corner, but also for a data set spanning multiple tracks as presented here.
Figure \ref{fig_overall:scores1} depicts the scores for pure brake and pure steer.  \\
In the following we exemplary compare the drivers D5 and D6. Under pure braking in \autoref{fig_overall:scores1}a, D6 has a higher total score \STot (Remark 1). We see that the difference arises because of a higher handling score \SHand, i.e. D6 is able to take the vehicle closer to its dynamic limit (Remark 2). Looking at  \SVeh, D5 has a slightly higher score, indicating that the optimised  vehicle state  is closer to the ideal tire state. To sum up,  the higher  \SVeh score cannot compensate the lower \SHand score. \\
 Under pure steer D5 has a lower \STot score, as shown in \autoref{fig_overall:scores1}b (Remark 3). Contrary to the difference in pure brake, the difference  originates from the \SVeh score (Remark 4), i.e. D5 has a less advantageous vehicle setup or driving line. The handling scores for both drivers are comparable. \\
In conclusion, D6 is overall better than D5 in pure brake and pure steer. The fact that D6 performed better than D5 in the analysed data set is supported by the better lap times and average speeds of D6 which are shown in \autoref{fig_track_metrics}a. However, in contrast to the lap time difference, the score analysis of this work allows much more detailed conclusions: D5 has to improve the handling skills under pure brake or needs a vehicle setup that has a better driveability.

\subsubsection{Amateur Drivers}
\begin{figure}
\begin{center}
\begin{subfigure}[c]{0.9\textwidth}
\begin{center}
\includegraphics[width=0.78\textwidth]{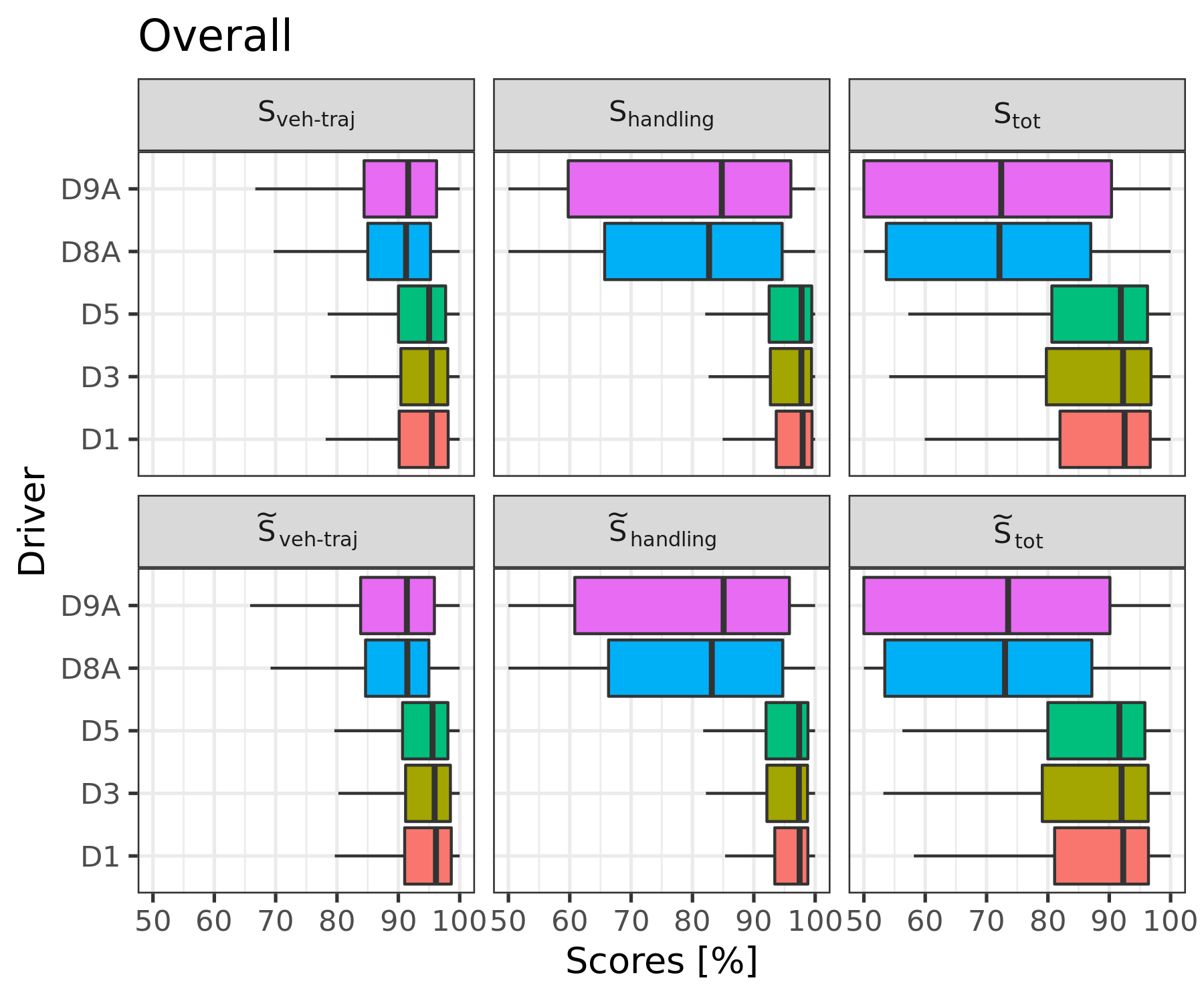}
\end{center}
\caption{Score overview for track T2 with amateur and professional drivers. In the first row are the score obtained by the optimisation procedure and in the second row the scores estimated by the machine learning predictor.  }
\end{subfigure}
\begin{subfigure}[c]{0.9\textwidth}
\begin{center}
\includegraphics[width=0.78\textwidth]{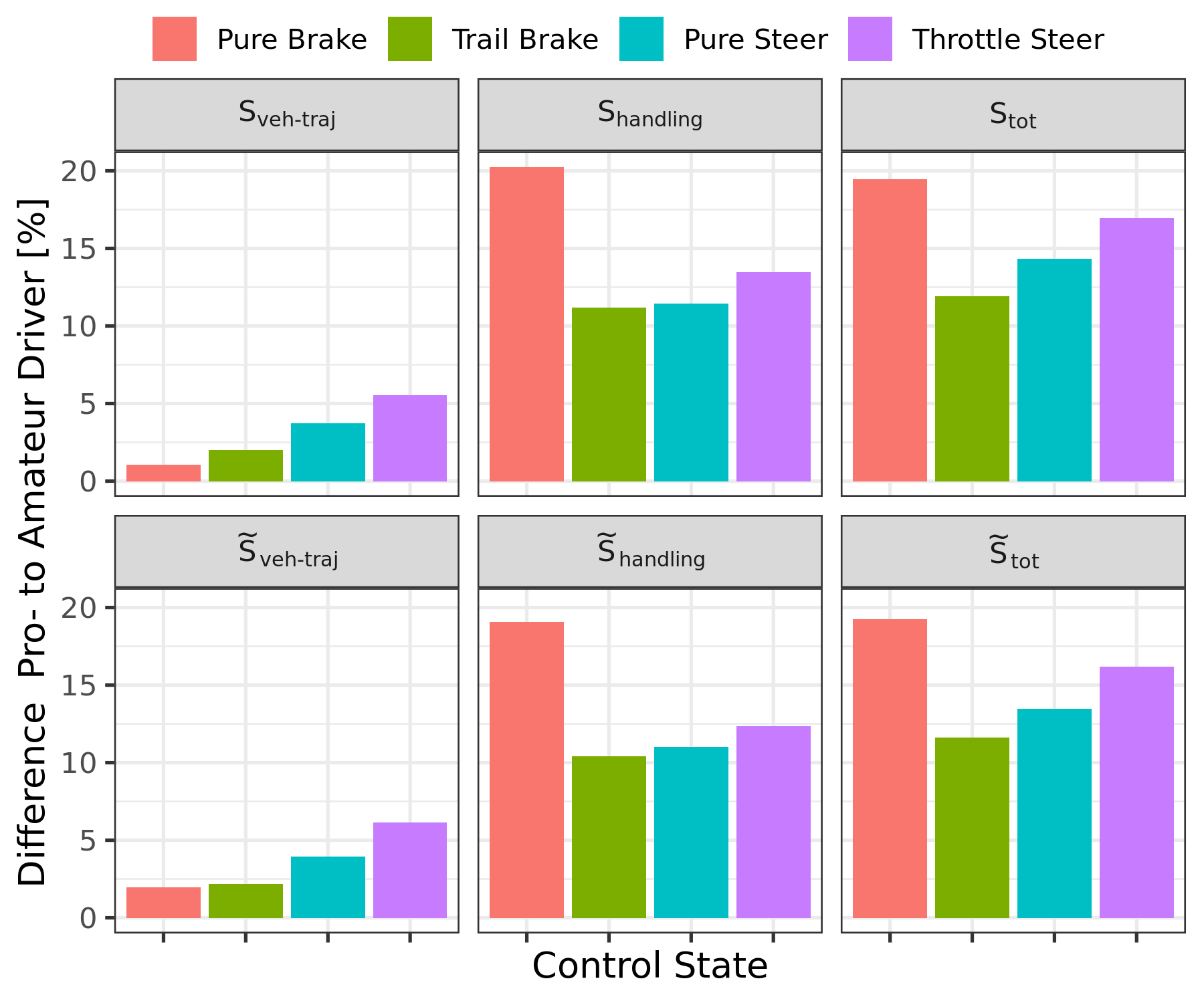}
\end{center}
\caption{Difference between professional and amateur drivers for the control states. As expected the professional drivers have higher scores in all cases, therefore the difference is always positive. The difference in \STot comes mainly from \SHand. In the second row are again the predictions of the machine learning model.}
\end{subfigure}
\end{center}
\caption{Professional versus amateur drivers. }
\label{fig_overall_bob}
\end{figure}

The differences in all of these scores were subtle for the professional drivers, which is natural given that they have been training for years to achieve the maximal performance.
To test our proposed method further and make it more interesting for a broader audience, we examine also amateur drivers with road car driving skills but no motorsport race experience. For that purpose, they drove in the same simulator environment as the professional drivers with a comparable vehicle setup on the track T2. In \autoref{fig_track_metrics}b significant differences between the two classes can be observed. For the amateurs, the lap time $t_\mathrm{lap}$ is 10 to $30\%$ higher and the average speed $v_\mathrm{avg}$ 10 to $20\%$ slower. The lap distance $s_\mathrm{lap}$ is  shorter on average, which is a consequence of the much lower speed, allowing smaller cornering radii. D9A achieved quicker lap times than D8A, however, D8A was more consistent, especially in lap distance. 
 Unsurprisingly,  significant differences in \STot are observed. Comparing the other two scores, the main difference comes from \SHand.  Since all drivers drove a comparable vehicle setup, the differences can be clearly attributed to the trajectory.\\
 Instead of providing the diagrams for all control states we summarized the information in \autoref{fig_overall_bob}b. The mean difference between the professional and amateur drivers is depicted for every control state. Most of the amateurs' performance is lost under braking. Almost all of that is due to a worse handling score. The amateur drivers are worse at controlling the vehicle and we observed that they did not exploit the aerodynamic  abilities of the vehicle. Because of high downforce, it is possible to brake much more at high speeds than at lower speeds. 
The highest difference in \SVeh is observed for throttle steer. Since the vehicle setups are equal, the reason has to be a difference in driving line. What is more, the difference in \SHand is the second highest leading to the highest \STot difference after pure brake. \\
To sum up, the professional drivers' driving line is superior to the amateurs. However, the difference is much larger for the  handling score. Most of the overall difference in \STot is a result of the much lower \SHand score and not so much \SVeh. For the amateur drivers to improve they would need to practice vehicle handling to be able to drive a higher speed along their chosen trajectory. 

\subsection{Machine learning predictor evaluation}
We already showed in Section \ref{chap_model_sel} and  \autoref{fig_lstm_res}b that the predictions of the machine learning model have a RMSE in the area of $2\%$ for \SHand and \STot and around $3\%$ for \SVeh on the considered data set with the professional drivers. Consequently, the predictions are in good accordance with the dynamic limit optimisation method while greatly decreasing computation time. This is supported by \autoref{fig_scores_example}b where the score distributions for reference and prediction are very similar for the whole data set.\\ 
The prediction also shows good results for the amateurs who were not part of the training set and whose driving styles were very different compared to the training data. In this case the RMSE is approximately $3\%$ for \SHand and \STot and  $5\%$ for \SVeh. 
Comparing the optimisation scores in the first row of \autoref{fig_overall_bob}b  and the predictions in the second row they are very similar. Interpretation and analysis of both methods would lead to  the same conclusion. What is more the difference between professional and amateur drivers is also very similar for both methods (\autoref{fig_overall_bob}c). \\
In terms of computational speed, the machine learning predictor needed $12$ seconds for inference (i.e. processing data through a trained neural network) on 1226 laps on a \emph{NVIDIA\textregistered Tesla\textregistered V100} GPU, which is a specialized deep learning graphics card. High-end consumer graphic cards will achieve a performance in the same order of magnitude. In contrast, the optimisation methods needed over $68$ hours for the same task on a  computer with two 10-core \emph{Intel\textregistered Xeon\textregistered Silver 4144} CPUs.  Due to the different hardware and architecture (machine learning on the GPU, optimisation the CPU), the comparison might not be completely fair, however, the results  show the massive improvement in computation time very clearly. \\
In conclusion, the  robustness of the machine learning model and the low prediction error allows to replace the computational expensive optimisation methods with the machine learning predictor for our specific use case. Given the very fast computation time, a real time execution is also possible. 

\section{Conclusion}
The presented methods showed their practical use for analysing drivers on a race track. However, it could be applied to all scenarios where the goal is to handle a vehicle at the limit. For road car driving, this can be the case for evasion manoeuvres or under slippery conditions. What is more, control systems for autonomous vehicles  could be judged by the presented scores in a simulated environment and compared to professional drivers.  
However, what sets professional race drivers apart is their ability to not only handle a vehicles in a stable environment, but also in the real world under changing conditions. As a next step, we therefore propose a simulator study whereby the amount of grip is varied continuously. Having at hand the here presented driver evaluation methods, it could be investigated how the participants react to the grip changes and how long the adaptation time will be until the best possible performance is reached. We hope that further insight on how human experts achieve such high performance at controlling a vehicle at the limit helps the development of advanced autonomous control systems in the future. 

\section*{Acknowledgement}
This work was generously supported by BMW AG.

\bibliographystyle{elsarticle-num-names}
\bibliography{jvs_bib}

\end{document}